%% file: arxiv4.tex
\documentclass[journal]{IEEEtran}

\usepackage[english]{babel}
\usepackage[utf8x]{inputenc}
\usepackage[T1]{fontenc}
\usepackage[doublespacing]{setspace}

\usepackage{amsmath, amssymb, amsthm, amsbsy, amsfonts,  cuted, colortbl, color, soul}
\usepackage[numbers]{natbib}
\usepackage{graphicx, comment,xcolor, enumitem, stfloats, setspace}
\usepackage{indentfirst} 
\usepackage[colorinlistoftodos]{todonotes}
\usepackage[colorlinks=true, allcolors=blue]{hyperref}

\allowdisplaybreaks

\theoremstyle{definition}
\newtheorem{thm}{Theorem}
\newtheorem{defn}{Definition}
\newtheorem{lem}{Lemma}
\newtheorem{cor}{Corollary}

\newtheorem{rem}{Remark}

\newcommand{\w}{\mathbf{w}}
\newcommand{\W}{\mathbf{W}}
\newcommand{\x}{\mathbf{x}}
\newcommand{\X}{\mathbf{X}}
\newcommand{\y}{\mathbf{y}}
\newcommand{\z}{\mathbf{z}}
\newcommand{\vv}{\mathbf{v}}
\newcommand{\aaa}{\mathbf{a}}
\newcommand{\h}{\mathbf{h}}
\newcommand{\f}{\mathbf{f}}
\newcommand{\e}{\mathbf{e}}
\newcommand{\F}{\mathbf{F}}
\newcommand{\E}{\mbox{E}}
\newcommand{\bb}{\mathbf{b}}
\newcommand{\N}{\mathbf{N}}

\newcommand{\1}{\mathbf{1}}
\newcommand{\0}{\mathbf{0}}
\newcommand{\tabitem}{~~\llap{\textbullet}~~}

\usepackage[]{algorithm2e}
\begin{document}
\title{Whiteout:  Gaussian Adaptive  Noise Injection Regularization in Deep Neural Networks}
\author{Yinan~Li and Fang~Liu
\thanks{\hspace{-12pt} Department of Applied and Computational Mathematics and Statistics}
\thanks{\hspace{-12pt} University of Notre Dame, Notre Dame, IN 46556, U.S.A.}
\thanks{\hspace{-12pt} email: fang.liu.131@nd.edu}
\thanks{\hspace{-12pt} We thank Mr. Ruoyi Xu for running the MNIST and CIFAR-10 experiments.}}

\setstretch{1.0}
\maketitle
\begin{abstract}	
Noise  injection (NI) is an efficient technique to mitigate over-fitting  in neural networks  (NNs). The Bernoulli NI procedure as implemented in dropout and shakeout has connections with $l_1$ and $l_2$ regularization for the NN model parameters. We propose whiteout, a family NI regularization techniques (NIRT) through injecting adaptive Gaussian noises during the training of NNs. Whiteout is the first NIRT than imposes a broad range of the $l_{\gamma}$ sparsity regularization $(\gamma\in(0,2))$ without having to involving the $l_2$ regularization. Whiteout can also be extended to offer regularizations similar to the adaptive lasso and group lasso.  We establish the regularization effect of whiteout in the framework of generalized linear models with closed-form penalty terms and show that whiteout stabilizes the training of NNs with decreased sensitivity to small perturbations in the input.  We establish that the noise-perturbed empirical loss function (pelf) with whiteout converges almost surely to the ideal loss function (ilf), and the minimizer of the pelf is consistent for the minimizer of the ilf. We derive the tail bound on the pelf to establish the practical feasibility in its minimization. The superiority of whiteout over the Bernoulli NIRTs, dropout and shakeout, in learning NNs with relatively small-sized training sets  and non-inferiority in large-sized training sets is demonstrated in both simulated and real-life data sets. This work represents the first in-depth theoretical, methodological, and practical examination of the regularization effects of both additive and multiplicative Gaussian NI in deep NNs.
\end{abstract}

\IEEEpeerreviewmaketitle

\section{Introduction}

Deep  NNs are prone to over-fitting given the large amounts of parameters involved in the multiplicity of layers and the large number of  nodes. Some of the earlier approaches to mitigate overfitting include unit pruning \citep{Ivakhnenko71, Mozer1989},  weight pruning \citep{LeCun1990}, $l_2$ regularization (weight decay)  \citep{Hanson1989},  max-norm \citep{maxnorm}, and early stopping \citep{Yu2005,Yuan2007},  among others.  There is also a resurgent of noise injection (NI) regularization techniques (NIRT), including dropout\citep{dropout12, dropout14}, dropconnect \citep{Wan2013}, maxout \citep{Goodfellow2013}, shakeout \citep{Kang2016}. In the rest of this section, we briefly discuss the existing NIRTs and then propose  whiteout, a family of NIRTs that not only impose similar regularization effects as some of the existing NIRTs, but also bring in new regularization types for mitigating overfitting in NNs in a more efficient manner than some  existing NIRTs in some settings.

\subsection{Existing work on NIRTs in NNs}\label{sec:related}
NI improves the generalization ability of a trained NN, especially in fully connected NN layers. Some of the early work of NIRTs in NNs  appeared  in the 1980's and 1990's. For example, \citet{Plaut86} and  \citet{Sietsma88} found experimentally that adding noise to the input during the training of a NN via the back-propagation had a remarkable effect on the generalization capability of the network. \citet{Holmstrom1992} examined feedforward NNs and suggested that injecting noises to training samples can be regarded as drawing samples from the kernel density estimation of the true density. However, they did not establish that NI improves the generalization error of a trained NN. \citet{NI92} formulated the NIRT in the input nodes as a way to decrease the sensitivity of a learned NN to small perturbations. \citet{Grandvalet1997} examined the theoretical properties and offered an explanation on the improved generalization of Gaussian NI (with a constant variance) to the input nodes via connecting NI with heat kernels. None of the work examined the regularization effects of NIRTs on NN parameters.

With the reactivation of deep learning since 2006 \citep{hinton06}, the research and application of NIRTs also enjoy a resurgence. The recent NIRTs, including dropout, dropconnect, standout, and shakeout, can all be viewed as injecting Bernoulli noises into a NN during  training. In dropout,  nodes in the input or the hidden layers are randomly dropped with some fixed probabilities  during the  training of a deep NN.  
Dropout has been shown to yield the $l_2$ regularization on the NN parameters in the framework of generalized linear models (GLMs) \citep{dropout12}.  Various  extensions to dropout have been also proposed. Maxout facilitates optimization  and improves the accuracy of dropout with  a new activation function \citep{Goodfellow2013}. Fast dropout speeds up the computation of dropout via a Gaussian approximation instead of randomly dropping nodes \citep{Wang2013}. Dropconnect applies  Bernoulli noises to weights instead of nodes \citep{Wan2013}.  Partial dropout/dropconnect combines weight decay, model averaging, and network pruning; and regularizes restricted Boltzmann machines (RBMs) \citep{partialdrop}.  Standout or adaptive dropout trains NNs jointly with a binary belief network that selectively sets nodes to zero \citep{Ba2013}. Shakeout extends the $l_2$ regularization in dropout by applying multiplicative adaptive Bernoulli noises to input and hidden nodes during training  to achieve  a combined  $l_1$ and $l_2$ regularization effect \citep{Kang2016}. The $l_1$ and $l_2$ regularization can also be realized by imposing the Laplace and Gaussian priors on NN parameters, respectively, in a variational Bayesian framework \citep{alex11}.  

Gaussian NI in NNs was briefly discussed in \citet{alex11} from a Bayesian perspective. \citet{dropout14}, motivated by the Bernoulli NI in dropoout,  suggested multiplicative Gaussian noises with mean 1 and a constant variance. They compared its performance with Bernoulli NI in the MNIST and CIFAR-10 data when the variances of the two types of noises were designed to the same, and found the Gaussian multiplicative NI was comparable or slightly better. However, they provided no theoretical exploration. \citet{Kang2016} demonstrated empirically that multiplicative Gaussian NI with a fixed variance  appeared to yield very similar performance as dropout in regularizing the weight parameters in NNs. In summary, existing research on Gaussian NIRTs is limited in scope and depth, and the methodological, theoretical, and empirical investigation lacks on how Gaussian NI can helps to regularize NNs and how it compares with Bernoulli NI. 

 \subsection{Our Contributions}\label{sec:contribution}
We propose and study \emph{whiteout}, a family of NIRTs, that injects adaptive Gaussian noises  during the training of a NN  (``adaptive'' in this context refers to that the variance term of the injected noises is a function of the weight parameters that changes  with the weight updates during training).  Our contributions are summarized below. 
\begin{enumerate}[leftmargin=12pt] \setlength\itemsep{0pt}
\item To the best of our knowledge, this is the first in-depth work that explores the regularization effects of Gaussian NIRTs from the methodological, theoretical, and empirical perspectives.
\item We propose whiteout as a \emph{family} of various NIRTs rather than a single NIRT. By setting the three tuning parameters, whiteout not only leads to the $l_2$  regularization and the $l_1$+ $l_2$  regularization,  but also brings in new NIRTs that connect with a wide range of regularizers commonly used in statistical training  of large regression models, such as lasso, bridge, adaptive lasso, and  group lasso. Whiteout is the first NIRT that can impose the $l_1$ constraint, and more broadly speaking, the $l_{\gamma}$ sparsity constraint ($\gamma \in(0,2))$,  without having to include the $l_2$ regularization component (the Bernoulli NIRT dropout leads to the $l_2$ regularization, while shakeout introduces the $l_1$+ $l_2$  regularization).
\item We show in both simulated and real-life experiments that whiteout outperforms Bernoulli NIRTs (e.g., dropout and shakeout) when the training size is small with its effectiveness in imposing flexible sparsity constraints on NN parameters. As such, whiteout can effectively discount noisy and irrelevant input and hidden features in prediction.
\item We establish whiteout as an effective  regularization technique from the following perspectives: first, whiteout is associated with a penalized objective function with a closed-from penalty term for model complexity in generalized linear models;  second, whiteout can  be regarded as a technique to improve robustness and decrease the sensitivity of a learned NN model to small perturbations in data.
\item We provide a thorough investigation of theoretical properties of the whiteout noise perturbed empirical loss function (pelf) and its almost sure (a.s.) convergence  toward the ideal loss function (ilf), and the consistency of the minimizer of the pelf to the minimizers of the ilf. The theoretical conclusions can also be extended to other NIRTs under the same regularity conditions. Though there existed previous work along the same lines (e.g., \citet{Grandvalet1997,Holmstrom1992}), they only examined NI in the input nodes and when noises are sampled from a distribution of fixed variances. To the best of our knowledge, this work the first on establishing the a.s. convergence of the pelf (to the ilf), the consistency of its minimizer, and its fluctuation (tail bound) around its expectation with respect to the distribution of  inject noises when NI occurs in both the input and hidden layers and when the variances of Gaussian noises change during training (aka ``adaptive'').
\item We develop a backpropagation (BP) procedure that easily incorporates the whiteout NI during training (App. \ref{sec:bp})
\item We show that whiteout can also be applied to unsupervised learning, such as RBMs  and auto-encoders (App. \ref{sec:unsupervised}).
\end{enumerate}
In what follows, we first introduce whiteout and demonstrate the selection of the tuning parameters (Section \ref{sec:whiteout}). We then establish the regularization effects of whiteout by formulating the noise injected optimization as a penalized likelihood problem in the context of generalized linear models (Section \ref{sec:regularization})  and show that whiteout stabilizes a learned NN with robustness to small external perturbation  (Section \ref{sec:sensitivity}). We illustrate the applications of whiteout in simulated and real-life data experiments and compare its prediction accuracy against dropout, shakeout, and no regularization (Sections \ref{sec:simulation} and \ref{sec:experiment}). We end the discussion in Section \ref{sec:discussion} with some final remarks.

\section{Whiteout}\label{sec:whiteout}
Gaussian NI can be either additive or multiplicative, both of which have appeared in the literature and both are assumed to have a constant variance. For completeness, we present both the additive and multiplicative whiteout noises (Sections \ref{sec:additive} and \ref{sec:multiplicative}). We compare their performance in prediction through experiments (Section \ref{sec:simulation} and \ref{sec:experiment}), and briefly discuss the pros and cons of each in Section \ref{sec:discussion}.

\subsection{Additive Noise in Whiteout}\label{sec:additive}
For whiteout with additive noise,  a noise term is drawn from the Gaussian distribution and then added to the original node value in the input or hidden layers. Let $l$ be the index for layers $(l=1,\ldots,L-1$, and $L$ is the output layer), and $j$ be the index for the nodes in layer $l$ ($j=1,\ldots, m^{(l)}$), and the weight connecting the $j$-th node in layer $l$ and the $k$-th node in layer $l+1$ is denoted by $w^{(l)}_{jk}$. The perturbed $j$-th node in layer $l$ $\tilde{X}_j^{(l)}$ is given by
\begin{align}
\tilde{X}_j^{(l)}&\!\!=\!X_j^{(l)}\!\!+\!e_{jk}, \mbox{where }e_{jk}\!\overset{\text{ind}}{\sim}\!N\!\left(0,\sigma^2|w^{(l)}_{jk}|^{-\gamma}\!\!+\!\lambda)\right).\!\!\label{eqn:additive}
\end{align}
The whiteout noise is \emph{adaptive} in the sense that its variance is a function of $w^{(l)}_{jk}$, which keeps being updated during training. $\sigma^2\ge0$, $\lambda\ge0$, and $\gamma\in(0,2)$  are  tuning parameters, the interpretation and selection of each are presented in Sec \ref{sec:tuningparameter}. Once these tuning parameters are chosen, they will be fixed throughout the training process.


\subsection{Interpretation and Specification of Tuning Parameters}\label{sec:tuningparameter}
The larger $\sigma^2$ and $\lambda$ are, the more regularization effects there will be on the weights (more weights being shrunk towards or set around 0). 
The ratio of $\sigma^2$ and $\lambda$ determines the relative regularization effects between sparsity and $l_2$ on the weight parameter in a NN. When $\sigma^2=0$, weights are regularized in the $l_2$ fashion. When $\lambda=0$, some type of $l_{2-\gamma}$ regularization (sparsity) will be imposed on weights, and what type depends the value of $\gamma$. For example, when $\gamma=1$, the sparsity regularization is $l_1$; when $\gamma \rightarrow 2$, the sparsity regularization approaches $l_0$.

In the practical implementation of whiteout, while all the tuning parameters could be chosen by the cross validation (CV), it can be challenging computationally given there are three of them. Instead, we recommend users first  specify values for one or two tuning parameters based on what type of regularization they would like to achieve when training a NN, and  then leave one or two tuning parameters to be selected by the CV. For example, setting $\sigma^2=\lambda$ would lead to a regularization somewhere between sparsity and $l_2$, and we then apply the CV to choose $\gamma$ and $\sigma^2=\lambda$ alternatively.  In addition, the magnitude of Gaussian noises should be kept below the fluctuation of the input; otherwise, it could result in underfitting and lead to biased predictions. The tuning of $\gamma$ controls the type of sparsity reguarlization $l_{2-\gamma}$. Our empirical studies  suggest that a good $\gamma$ can often be found in the neighborhood of $(0.5, 1.5)$. $\gamma\rightarrow2$ could lead to poor predictions due to large injected noises; and $\gamma\rightarrow0$ yields minimal sparsity regularization on top of $l_2$. If users prefer not to tune $\gamma$ to save on computational cost, $\gamma$ may be fixed at 1 without significantly affecting the prediction accuracy.

Listed in Table \ref{tab:whiteout} are some examples of the whiteout noise by setting the tuning parameters at some specific values (for notation simplicity, $w_{jk}$ is used in place of $w^{(l)}_{jk}$). These examples also help to better understand the functionality of each tuning parameter. The names of the noise types are
\begin{table}[!htp]
\centering\resizebox{\linewidth}{!}{
\begin{tabular}{lll}
\hline
\multicolumn{3}{c}{\cellcolor{gray!20} whiteout $e_{jk}\sim N(0,\sigma^2|w_{jk}|^{-\gamma}\!\!+\!\lambda)$}\\
\hline
tuning & regularization & noise name \\
parameter  & effect  & \& distribution \\
\hline
$\lambda=0$ & bridge $l_{2-\gamma}$ \citep{bridge} &  Gaussian bridge  (gab) \\
 &&$e_{jk}\sim N\left(0,\sigma^2|w_{jk}|^{-\gamma}\right)$\\
$\lambda=0, \gamma=1$ & lasso $l_1$ \citep{lasso96}   & Gaussian lasso  (gala)  \\
&& $e_{jk}\sim N\left(0,\sigma^2|w_{jk}|^{-1}\right)$\\
$\sigma^2=0$ & ridge $l_2$   & Gaussian ridge  (gar)\\
&& $e_{jk}\sim N (0,\lambda )$\\
 $\gamma=1$  &  elastic net (EN)  &   Gaussian EN  (gen) \\
 &$l_1+l_2$  \citep{en05} & $e_{jk}\sim N (0,\lambda )$\\
 \hline
\end{tabular}}
\centering\resizebox{\linewidth}{!}{
\begin{tabular}{ll}
\hline
\multicolumn{2}{c}{\cellcolor{gray!20} extensions of whiteout (see the text below the table)}\\
\hline
regularization effect & noise distribution \\
\hline
adaptive lasso \citep{adaptivelasso06} & Gaussian adaptive lasso  (gaala) \\
& $e_{jk}\sim N\left(0,\sigma^2(|w_{jk}|)^{-1}|\hat{w}_{jk}|^{-\gamma}\right)$\\
group lasso \citep{glasso06} & Gaussian group  (gag) \\
& $e_{jk(g)}\sim N\left(0, \sigma^2(\w'_g\mathbf{K}_g\w_g)^{1/2}(p_g w_{jk}^2)^{-1}\right)$\\
& for groups $g=1,\ldots,G$\\
\hline
\end{tabular}}
\caption{Example of whiteout noise types and whiteout extensions and their regularization effects}\label{tab:whiteout}
\end{table}
motivated by the regularization effect that each brings in the context of GLMs (detailed in Section \ref{sec:regularization}). The table also lists some extensions to the basic whiteout noise type to yield more  types of regularization effects, such as the adaptive lasso and the group lasso. The adaptive lasso was introduce in the regression setting and is an improvement over the lasso with its  oracle properties \citep{adaptivelasso06}, and the group lasso was introduced  to allow predefined groups of attributes to be selected into or out of a regression model together \citep{glasso06}. In the setting whiteout noise,  $\hat{w}_{ij}$ in the gaala noise  is a weight estimate, say as learned from a deep learning algorithm without regularization 
The gag whiteout noise can be applied to penalize predefined groups of input nodes. The number of groups in the input nodes is denoted by $G$, and the size of group $g$ by $p_g$  for  $g=1,\ldots,G$, and $\w_g$ contains all the weights associated with the input nodes $X_{j}$ from group $g$ ($j=1,\ldots, p_g$), and $\mathbf{K}_g$ is a positive-definite matrix. When $p_1 =\ldots=p_G =1$ (one node per group), the gag noise  reduces to the gala noise. While the gab, gen and gaala noises can be injected to both input and hidden nodes, the gag noise makes the most sense in perturbing the input nodes since grouping of hidden nodes, which represent abstract features that do not necessarily have any physical meanings, are hard to justify.

\subsection{Multiplicative Noise in Whiteout}\label{sec:multiplicative}
In addition to the additive  noise, whiteout can also inject multiplicative noises in a NN. Using the same notations as in Section \ref{sec:additive},
\begin{align}
&\tilde{X}_j^{(l)}\! =\!X_j^{(l)}\epsilon_{jk}, \mbox{where } \epsilon_{jk}\!\overset{\text{ind}}{\sim}\!N\left(1, \sigma^2|w^{(l)}_{jk}|^{-\gamma}\!+\!\lambda\right)\label{eqn:multiplicative}\\
&=\!X_j^{(l)}\!\!+\!e'_{jk}, \mbox{where }e'_{jk}\!\overset{\text{ind}}{\sim}\!N\!\!\left(\!0,\left(X_j^{(l)}\right)^{\!2}\!\!\left(\!\sigma^2|w^{(l)}_{jk}|^{-\gamma}\!\!+\!\lambda\right)\!\!\right)\!\!\! \label{eqn:multiplicative1}
\end{align}
Though the multiplicative noise can be re-expressed in terms of an additive noise (Eqn \ref{eqn:multiplicative1}), the dispersion of the reformulated additive noise depends on the node value $X_j^{(l)}$. This implies extreme node values could be generated if $|X_j^{(l)}|$ is already large in magnitude (in contrast, the dispersion of additive whiteout noise is independent of the node values). Large noises can lead to harsher penalty on the weights connected with large nodes. When the nodes are somewhat ``outlying'', the weights connected to it being  harshly penalized might lead to additional robustness effects in the sense that it may help to reduce the network's sensitivity to outlying nodes (more research is needed to confirm whether the conjecture is true).

\section{Justification of Whiteout for model regularization with improved generalization}\label{sec:why}
In this section, we justify whiteout as a NIRT that improves the generalization of a learned NN from two perspectives. Section \ref{sec:regularization} connects whiteout with various regularization effects on model parameters in the setting of GLMs; and Section \ref{sec:sensitivity} shows that whiteout stabilizes and robustifies a learned NN in that the learned NN through whiteout offers low sensitivity to small perturbation in the input data.

\subsection{regularization effects of whiteout}\label{sec:regularization}
Whiteout injects independent additive or multiplicative noises into  input and hidden nodes in a NN.   A common framework where NI  is established as a regularization technique is the GLMs based on the exponential family \citep{Kang2016, An1996, Bishop1995,  wager13}. In a GLM,  the conditional distribution of output $Y$ given inputs $\mathbf{X}\in \mathcal{R}^p$ is modeled as
\begin{equation}\label{eqn:exp2}
f(Y|\X,\mathbf{w})=h(Y,\tau)\exp\left((\boldsymbol{\eta}\mathbf{T}(Y)-A(\boldsymbol{\eta}))/d(\tau)\right),
\end{equation}
where $\boldsymbol{\eta}=\X\w$  is the natural parameters, $\w$ refers to the regression coefficients associated with $\X$,  and $\tau$ is the dispersion parameter. The functional forms of $h(\cdot,\cdot),\mathbf{T}(\cdot)$ and $A(\cdot)$ are known given an assumed distribution for $Y$. For example, if $Y$ is Gaussian with a constant variance, then $d(\tau)=\sigma^2, \mathbf{T}(Y)=Y, A(\boldsymbol{\eta})=\boldsymbol{\eta}^2/2$ and $h(Y,\tau)=\exp(-Y^2/(2\sigma^2))/\sqrt{2\pi\sigma^2}$. If $Y$ is Bernoulli, then  $d(\tau)=1, \mathbf{T}(Y)=Y, A(\boldsymbol{\eta})=\log(1-e^{\eta}/(1+e^{\eta}))$ and $h(Y,\tau)=1$.

The negative log-likelihood for the model given independent training cases $(\mathbf{x}_i,y_i)$ for $i=1,\cdots,n$ is
\begin{equation}\label{eqn:mle}
\!\!\!l(\mathbf{w|\x},\y)\!=\!\textstyle\sum_{i=1}^n (-\boldsymbol{\eta}\mathbf{T}(y_i)\!+\!A(\boldsymbol{\eta_i}))/d(\tau)\!-\!\mbox{log}(h(y_i,\tau)).
\end{equation}
Whiteout substitutes the observed $\x_i$  in Eqn (\ref{eqn:mle}) with its noise-perturbed version $\tilde{\x}_{i}$ defined in Eqns (\ref{eqn:additive}) or (\ref{eqn:multiplicative}). The noise  perturbed negative log-likelihood  is
\begin{equation}\label{eqn:mle.p}
l_p(\mathbf{w|\tilde{\x}},\y)=\textstyle \sum_{i=1}^n l(\mathbf{w}|\tilde{\x}_i,y_i).
\end{equation}
Lemma \ref{lem:Rw} below establishes that the expected $l_p(\mathbf{w|\tilde{\x}},\y)$ over the distribution of injected noises is a penalized likelihood with the raw data with a regularization term  $R(\w)$.
\begin{lem}[\textbf{Penalized likelihood in GLMs with whiteout}]\label{lem:Rw}
The expectation of Eqn (\ref{eqn:mle.p}) over the distribution of noise is
\begin{align}
&\mbox{E}_{\e}(\textstyle\sum_{i=1}^{n}l_p(\w|\tilde{\x}_i,y_i))\!=\!\textstyle\sum_{i=1}^{n}\!l(\w|\x_i,y_i)\!+\!\frac{R(\w)}{d(\tau)},\!\label{eqn:regloss}\\
&\mbox{where } R(\w)\triangleq\textstyle\sum_{i=1}^{n}\mbox{E}_{\e}(A(\tilde{\x}_i\w))-A(\boldsymbol{\eta}_i)\notag\\
&\qquad\qquad\quad \approx \textstyle\frac{1}{2} \sum_{i=1}^{n}A''(\boldsymbol{\eta}_i)\mbox{Var}(\tilde{\x}_i\w).\notag
\end{align}
\end{lem}
\noindent The proof of  Lemma \ref{lem:Rw} is given in Appendix \ref{app:Rw}. Note that $A(\boldsymbol{\eta}_i)=A(\x_i\w)$  is convex and smooth in $\w$ \citep{MJ} in GLMs, and $R(\w)$ is always positive per the Jensen’s inequality \citep{wager13}.

Based on Lemma \ref{lem:Rw}, we examine the actual forms that the regularization term $R(\w)/d(\tau)$ in Eqn (\ref{eqn:regloss}) takes given some specific values of the tuning parameters $(\sigma^2, \lambda, \gamma)$. The results  for the additive noise are given in Theorem \ref{thm:regularization} and those for the multiplicative noise case are given in Corollary \ref{cor:multi.regularization}. 
\begin{thm}[\textbf{Regularization on $\w$ with additive whiteout noise}]\label{thm:regularization}
Let $\mathbf{\Lambda(\w)}\! =\!\mbox{diag}(\!A''(\x_1\w),\cdots\!,A''(\x_n\w))$ in the framework of GLMs.
\begin{itemize}
\item[a). ] whiteout with the additive gab noise leads to
\begin{equation}\label{eqn:add.gaalan}
R(\w)\approx(\sigma^2/2)\mathbf{1}^{T}\mathbf{\Lambda(\w)1}\big|\big||\w|^{2-\gamma}\big|\big|_1,
\end{equation}
where $\mathbf{1}_{n\times1}$ is a column vector of 1. The penalty $\big|\big||\w|^{2-\gamma}\big|\big|_1$ on $\w$ is similar to the bridge penalization \citep{bridge}, which reduces to the $l_1$ (lasso) penalty \citep{lasso96} when $\gamma=1$,  and to the $l_2$ (ridge) penalty when $\gamma=0$.
\item[b).] whiteout with the additive gen noise leads to
\begin{equation}\label{eqn:add.genn}
R(\w)\!\approx\!(1/2)\mathbf{1}^{T}\mathbf{\Lambda(\w)1}\left(\sigma^2||\w||_1\!+\! \lambda||\w||_2^{2}\right),
\end{equation}
which contains a similar norm on $\w$ as the EN ($l_1$ and $l_2$) regularization \citep{en05}.
\item[c).]  whiteout with the additive gaala noise leads to
\begin{align}\label{eqn:add.gaala}
R(\w)\approx(\sigma^2/2)&\mathbf{1}^{T}\!\mathbf{\Lambda(\w)1}\big|\big||\w||\hat{\w}|^{-\gamma}\big|\big|_1,
\end{align}
which contains a similar norm on $\w$ as the adaptive lasso regularization \citep{adaptivelasso06}.
\item[d).] whiteout with the additive gag noise  leads to
\begin{align}\label{eqn:add.gag}
\!\!\!\!\!\!R(\w)\!\approx\!\frac{\sigma^2}{2}&\mathbf{1}^{T}\!\!\mathbf{\Lambda(\w)1}\! \left(\textstyle\sum_{g=1}^{G}\!\big|\big|(\w'_g\mathbf{K}_g\w_g)^{\frac{1}{2}}p_g^{-1}
\! \big|\big|\right)\!,\!\!
\end{align}
which contains a similar norm on $\w$ as the group lasso penalization \citep{glasso06}.
\end{itemize}
\end{thm}
In addition to the various norms on $\w$, the penalty terms in Eqns (\ref{eqn:add.gaalan}) to  (\ref{eqn:add.gag}) also involve $\mathbf{\Lambda(\w)}$.  When $A''(\eta_i)$ does not depend on $\eta_i$ (thus $\w$), $R(\w)$  leads to  the nominal regularization.  For example, in linear models with Gaussian outcomes of constant variances,  $A(\eta_i)=\eta_i^2/2, A''(\eta_i)=1$, $\mathbf{\Lambda(\w)}=I_n$, 
and $R(\w)\approx \textstyle\frac{\sigma^2}{2}\sum_{i=1}^{n}\sum_{j=1}^{p}|w_j|^{2-\gamma}
=\frac{n\sigma^2}{2}\big|\big||\w|^{2-\gamma}\big|\big|_1$ in Eqn (\ref{eqn:add.gaalan}). If $A''(\eta_i)$ depends on $\eta_i$ (thus $\w$), the regularization effects on $\w$ through $R(\w)$  are not ``exact'' as the names suggest due to the  scaling of $\mathbf{\Lambda(\w)}$ on the norms of $\w$.  For example,  in logistic regression with binary outcomes, $A(\eta_i)=\ln(1+e^{\eta_i}), A''(\eta_i)=p_i(\w)(1-p_i(\w))$ with $p_i(\w)=\Pr(y_i=1|\x_i)=(1+\exp(-\x_i\w))^{-1}$,
and $R(\w)\approx \textstyle
\frac{\sigma^2}{2}\left(\sum_{i=1}^{n}p_i(\w)(1-p_i(\w))\right)\big|\big||\w|^{2-\gamma}\big|\big|_1$ -- the norm of $\w$ is scaled by the total variance of the binary outcome.
\begin{cor}[\textbf{Regularization on $\w$ with multiplicative whiteout noise}]\label{cor:multi.regularization}
Define $\Gamma(\w)\triangleq\mbox{diag}(\mathbf{x^{T}\Lambda(\w)x})$ in GLMs. 
	\begin{itemize}
		\item[a). ]  whiteout with the multiplicative gab noise leads to the bridge penalty term (that includes the $l_1$ and $l_2$ regularization as special cases)
		\begin{equation}\label{eqn:multi.gaalan}
		R(\w)\approx(\sigma^2/2)\big|\big|\Gamma(\w)|\w|^{2-\gamma}\big|\big|_1.
    \end{equation}
		\item[b). ]  whiteout with the multiplicative gen noise leads to the  $l_1$ + $l_2$ penalty term
		\begin{equation}\label{eqn:multi.genn}
		\!\!R(\w)\!\approx\!(\sigma^2/2)\big|\big|\Gamma(\w)|\w|\big|\big|_1\!\!+\!(\lambda/2)\big|\big|\Gamma(\w)|\w|^{2}\big|\big|_1.
    \end{equation}
        \item[c).]  whiteout with the multiplicative gag noise leads to the adaptive lasso penalty term
        \begin{equation}
        R(\w)\approx(\sigma^2/2)\big|\big|\Gamma(\w)|\w||\hat{\w}|^{-\gamma}\big|\big|_1.
    \end{equation}
		\item[d). ]  whiteout with the multiplicative gag noise leads to the group lasso penalty term
\begin{equation}\label{eqn:multi.gag}
\!\!\!\!\!\!\!\!R(\w)\!\approx\!(\sigma^2/2)\! \textstyle\sum_{g=1}^{G}\!\big|\big|\Gamma_g(\w)\big|\big|(\w'_g\mathbf{K}_g\w_g)^{\frac{1}{2}}\big|\big|p^{-1}_g\big|\big|_1,\!\!
\end{equation}
where $\Gamma_g(\w)$ is the sub-matrix of $\Gamma(\w)$ corresponding to $\w_g$.
\end{itemize}
\end{cor}
In the penalty terms in Eqns (\ref{eqn:multi.gaalan}) to  (\ref{eqn:multi.gag}), the norms are on $\Gamma(\w)\w$, a scaled version of $\w$,  rather than on $\w$ directly.  Plugging in the MLE $\hat{\w}$, $n^{-1}\x^{T}\Lambda(\hat{\w})\x=n^{-1}\sum_{i=1}^{n}\nabla^2l(\w^*|\x_i,y_i)$, which is an estimator of the Fisher information matrix in GLMs.  Since $\Gamma(\w)\triangleq\mbox{diag}(\mathbf{x^{T}\Lambda(\w)x})$ per definition,  whiteout with the  multiplicative noise can thus be regarded as regularizing $\w$ after scaling it with the diagonal Fisher information matrix. Dropout has a similar interpretation with the Bernoulli NI \citep{wager13}.

\subsection{Stabilization of Learned NNs  via Whiteout}\label{sec:sensitivity}
In this section, we establish that the whiteout procedure can stabilize and robustify a learned NN in the sense that the learning the NN  through whiteout NI takes into the sensitivity of the learned NN to small external perturbation in the input data.  Theorem \ref{thm:sensitivity} provides another justification to the generalization ability of whiteout in training NNs from a different perspective than that examined in Section \ref{sec:regularization}. 

Denote the training data  by $\z_i=(\x_i,\y_i)$, where $\x_i\!=\!(x_{i1},\cdots,x_{ip})$ for $i\!=\!1,\ldots,n$ with $p$ input nodes and $q$ output nodes, and the NN model by $\y_i=\mathbf{f}(\x_i|\w,\bb)$. Let $\mathbf{d}$  denote the independent small external perturbations to the $p$ input nodes $\x_i$, where $\E(d_{ij}\!)\!\!=\!0$ and $\mbox{V}(d_{ij}\!)\!\!=\!\varpi^2$ for $j=1,\ldots,p$.  Denote the predicted outcome given the unperturbed $\x_i$ from the learned NN via the additive whiteout noise $e_{jk}\sim$ N$\left(0,\sigma^2|w_{jk}|^{-\gamma}+\lambda\right)$ (Eqn \ref{eqn:additive}), injected into the input and hidden nodes by $\hat{\y}_i$, that from the trained NN without NI by $\bar{\y}_i$, and that given the externally perturbed input from the NN learned with whiteout NI by $\hat{\hat{\y}}_i$.
\begin{thm}[\textbf{Low sensitivity of leaned NN with whiteout}]\label{thm:sensitivity}
The expected value of the noise perturbed loss function over the distribution of the injected whiteout noise $\e^*$, $l_p(\w,\bb|\e^*,\x,\y)\!\!=\!\!\sum_{i=1}^{n}\!|\y_i\!-\!\hat{\y}_i|^{2}$ ($|\cdot|$ denotes the Euclidean norm), is approximately equivalent to  the sum of the original loss function $l(\w,\bb|\x,\y)\!=\!\sum_{i=1}^{n}\!|\y_i\!-\!\bar{\y}_i|^2$  and the sensitivity $ S(\w,\bb)$ of the NN,
\begin{equation}\label{eqn:sensivity}
\E_{\e^*}(l(\w,\bb|\e^*,\x,\y))\approx l(\w,\bb|\x,\y)+ a S(\w,\bb).
\end{equation}
$a >0$ is a tuning parameter and the sensitivity is defined as
\begin{align*}
S(\w,\bb)&=\sum_{i=1}^{n}\!\frac{\mbox{V}_{\e_i,\mathbf{d}_i}(|\boldsymbol{\Delta}_i|)}{\mbox{V}_{\mathbf{d}_i}(|\mathbf{d}_i|)}\\
&=p^{-1}\sum_{i=1}^{n}\!\sum_{q'=1}^q\!{\Psi}_{q',i}\!
\begin{pmatrix}
\!R\!+\!D_1\!\!&\!\!\! 0\!\!\\
\!0\!\!&\!\!\!D_{q',2}\!\!\\
\end{pmatrix}
\!{\Psi}_{q',i}^T\\
&=\textstyle p^{-1}\sum_{i=1}^{n}\!\sum_{q'=1}^q\!{\Psi}_{q',i}D{\Psi}_{q',i}^T,
\end{align*}
where $\boldsymbol{\Delta}_i\!=\!\hat{\hat{\y}}_i\!-\!\hat{\y}_i$, the difference between the predicted outcome through the NN learned with whiteout noise given externally perturbed and that given the unperturbed input, ${\Psi}_{i,q'}(\w,\bb)=\left(
\frac{\partial \f_{q'}^{(L-1):1}}{\partial{f}_{q',1}^{(1)}} \frac{\partial{f}_{q',1}^{(1)}}{\partial{\x}_i},\ldots,
\frac{\partial \f_{q'}^{(L-1):1}}{\partial{f}_{q',m^{(2)}}^{(1)}} \frac{\partial{f}_{q',m^{(2)}}^{(1)}}{\partial{\x}_i},
\frac{\partial\f_{q'}^{(L-1):2}}{\partial{f}_1^{(2)}}
\frac{\partial{f}_1^{(2)}}{\partial{\h}^{(2)}_i},
\ldots,\right.\\
\left.\frac{\partial f_{q'}^{(L-1):2}}{\partial {f}_{q',m^{(3)}}^{(2)}}
\frac{\partial{f}_{q',m^{(3)}}^{(2)}}{\partial{\h}^{(2)}_i},
\ldots,
\frac{\partial f_{q'}^{(L-1)}}{\partial{\h}^{(L-1)}_i},
\ldots,
\frac{\partial f_{q'}^{(L-1)}}{\partial{\h}^{(L-1)}_i}\right)$ is the gradient of $\mathbf{f}_{q'}$ for the $q'$-th output with regard to the perturbed and injected noises, $\f_{q'}^{l_1:l_2}$ is the compound function over layers $l_1$ to $l_2$ for the $q'$-th output node, $m^{(l)}$ is the number of nodes in layer $l$ $\left(m^{(1)}=p\right)$, $\h^{(l)}_i$ refers to the hidden nodes in layer $l$,
$\frac{\partial{f}_{q',j}^{(1)}}{\partial {\x}_i}\!=\!\left(\!\frac{\partial {f}_{q',j}^{(1)}}{\partial {x}_{i1}},\!\cdots,\!\frac{\partial {f}_{q',j}^{(1)}}{\partial {x}_{ip}}\!\right)^T$ for $j=1,\ldots,m^{(2)}$, and $\frac{\partial {f}_{q',j}^{(l)}}{\partial {\h}^{(l)}_i} \!=\!\left(\!
\frac{\partial {f}_{q',j}^{(l)}}{\partial {h}^{(l)}_{i1}},\!\cdots,\!
\frac{\partial {f}_{q',j}^{(l)}}{\partial {h}^{(l)}_{i,m^{(l)}}}\!\right)^T$ for $j=1,\ldots,m^{(l+1)}$ and $l=2,\ldots,L-1$.  $R$ is a symmetric band matrix  that captures the correlation among the injected and perturbed noise terms in the input nodes due to the shared $\mathbf{d}_i$ and $R[i,i\!+\!p]\!=\!1$ for $i= 1,\ldots,p(m-1)$ and 0 otherwise in its upper triangle,
$D_1\!=\!\mbox{diag}\!\left(\!\sigma^2\varpi^{-2}\big|w_{jk}^{(1)}\big|^{-\gamma}\!\!\!+\!\lambda\varpi^{-2}\!+\!1\right)$ for $j\!=\!1,\ldots,p$ and \\$k= 1,\ldots,m^{(2)}$,
$D_{q',2}\!=\!\mbox{diag}\!\left(D^{(2)},\ldots,D^{(L-2)},D_{q'}^{(L-1)}\right)$  \\ with $D^{(l)}\!=\!\mbox{diag}\!\left(\!\sigma^2\varpi^{-2}\big|w_{jk}^{(l)}\big|^{-\gamma}\!\!\!+\!\lambda\varpi^{-2}\!\right)$ for $l=2,\ldots, L-2$ \\ and
$D_{q'}^{(L-1)}=\mbox{diag}\left(\sigma^2\varpi^{-2}\big|w_{jq'}^{(L-1)}\big|^{-\gamma}\!\!\!+\!\lambda\varpi^{-2}\right)$.

The injected whiteout noises $\e^*$ leads to Eqn (\ref{eqn:sensivity}) is $\e_{i}^*=\e_{i}^{a}\!+\!\e_{i}^{b}$, where
$e_{ijk}^{a}\!\sim N(0,\sigma^{*2}\big|w_{jk}^{(1)}\big|^{-\gamma}+\lambda^{*})$ and $e_{ij}^{b}\!\sim N(0,p^{-1})$  for  input nodes; and $e_{ijk}^{a}\!\sim N(0,\sigma^{*2}\big|w_{jk}^{(l)}\big|^{-\gamma}+\lambda^{*})$ and $e_{ijk}^{b}= 0$ for hidden nodes in layer $l=2,\ldots,L-1$, with $\sigma^{*2}= a\sigma^2\varpi^{-2}/p$ and $\lambda^{*}=a\lambda\varpi^{-2}/p$ (implying the variance of whiteout noises $\e^*$ is proportional to $a$).
\end{thm}
\noindent The proof of Theorem \ref{thm:sensitivity} is given in Appendix \ref{app:sensitivity}.   Eqn (\ref{eqn:sensivity}) suggests minimizing the original loss function with a penalty term for the instability (sensitivity) of the network is approximately equivalent to minimizing the perturbed loss function with whiteout noise $\e^*$.  When the tuning parameter $a$ in Eqn (\ref{eqn:sensivity}) $\!\rightarrow\!0$ (i.e., minimal whiteout NI), the sensitivity of the learned NN can be undesirably large without being penalized for its instability. As $a$ increases (i.e., increased amounts of whiteout NI), the learned NN has to become more stable to yield small $S(\w,b)$ to maintain a small value for the sum of the original loss function and  $aS(\w,b)$. In other words, we would find a NN that minimizes the sum of the original loss function and the sensitivity of the NN by tuning $a$.
\section{Asymptotic Properties of Loss Function and Parameter Estimates in Whiteout}\label{sec:asymptotics}
We have shown in Sections \ref{sec:regularization} and \ref{sec:sensitivity} that whiteout is a  family of NIRTs that mitigates over-fitting and improves the robustness and generalization of a learned NN model. In this section, we examine the asymptotic properties of  noise-perturbed empirical loss functions with whiteout  and the estimates of NN parameters from minimizing the perturbed loss function. The goal is to establish that the minimizer of the perturbed loss function is consistent for minimizer of the loss function if the distributions of $\X$ and $\mathbf{Y}$ were known (as $n\rightarrow \infty$) and the epoch number $k\rightarrow\infty$ in a NN learning algorithm (e.g., back-propagation). We also investigate the tail bound on the noise-perturbed  empirical loss functions with a finite $k$ to establish the whiteout noise perturbed empirical loss function is trainable, which is important from the practical implementation perspective.

\begin{figure}[!htp]
\begin{center}
\includegraphics[width=\linewidth]{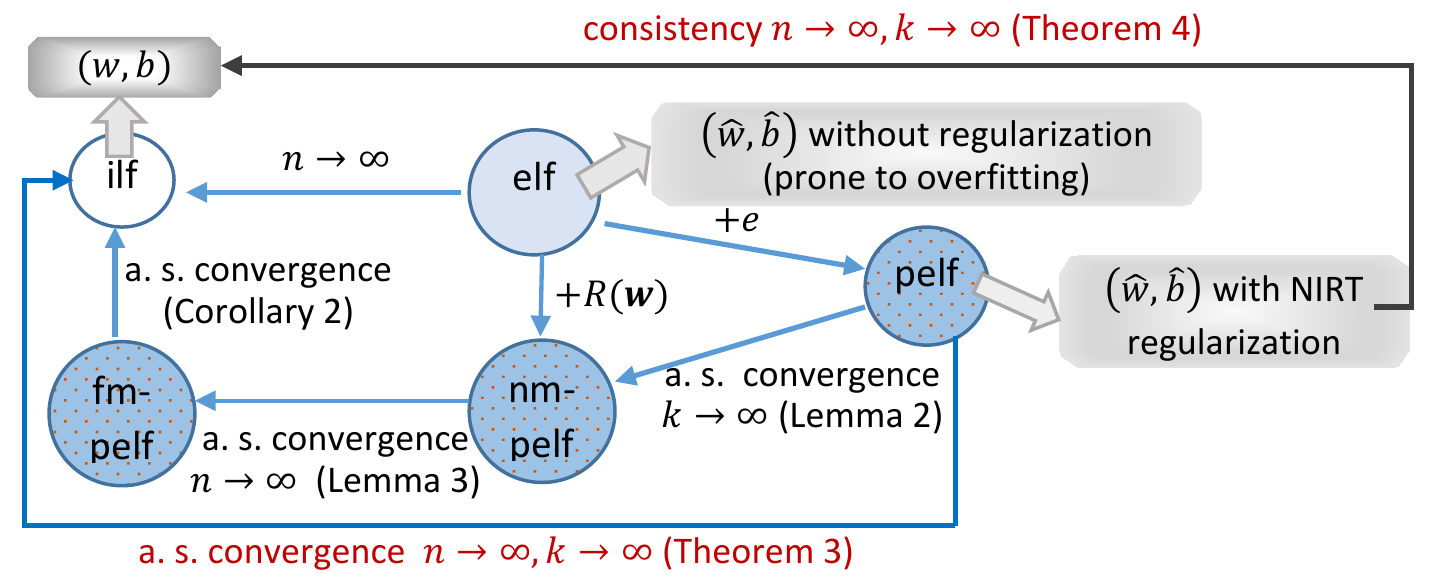}
\caption{Relationship among the loss functions}\label{fig:loss}
\end{center}
\end{figure}

Before we present the main results, it is important to differentiate among several types of loss functions. Understanding the differences among these loss functions facilitates the investigation of theoretical properties in NIRTs in general 
The definitions are general enough to take any form (e.g., the $l_p$ loss), though the $l_2$ loss is used in Definition \ref{def:loss}.
\begin{defn}[\textbf{Loss Functions in NIRTs}]\label{def:loss}
Let $p(\mathbf{X},\mathbf{Y})$ denote the unknown underlying distribution of $(\mathbf{X},\mathbf{Y})$ from which training data $(\x,\y)$ are sampled. Let  $f(\mathbf{Y}|\X,\w,\bb)$ be the composition of activation functions among the layers with bias and weight parameters  $\bb$ and $\w$ in a NN.
\begin{itemize}\item[]
\item[a). ] The \emph{ideal loss function (ilf)} is  $l(\w,\bb)=\E_{\x,\mathbf{y}}|f(\x|\w,\bb)-\mathbf{y}|^2$. $l(\w,\bb)$ is not computable since $p(\x,\mathbf{y})$ is unknown . 
\item[b). ] The \emph{empirical loss function (elf)} is $l(\w,\bb|\x,\y)=n^{-1}\sum_{i=1}^{n}|f(\x_{i}|\w,\bb)-\y_{i}|^{2}$.  $l(\w,\bb|\x,\y) \rightarrow l(\w,\bb)$ as $n\rightarrow\infty$.
\item[c). ] The \emph{noise perturbed empirical loss function (pelf)}   is $l_{p}(\w, \bb|\x,\y,\e)\!=\!(kn)^{-1}\!\sum_{j=1}^{k}\!\sum_{i=1}^{n}\!|f(\x_{i}, \e_{ij}|\w,\bb)-\y_{i}|^{2}$, where $\e_{ij}$ represents the collective noise injected into case $i$ in the $j\textsuperscript{th}$ epoch during training.
\item[d).]  The \emph{noise-marginalized perturbed  empirical loss function  (nm-pelf)}  is  the expectation of pelf over the
distribution of noise: $l_{p}(\w, \bb|\x,\y)=\E_{\e}(l_p(\w, \bb|\x,\y,\e))$. The nm-pelf can be interpreted as training a NN model by minimizing the perturbed empirical loss function with a finite $n$ and an infinite number of	epochs ($k\rightarrow\infty$).
\item[e).]  The \emph{fully marginalized perturbed empirical loss function (fm-pelf)}  is the expectation of nm-pelf over the distribution $p(\mathbf{x},\mathbf{y})$: $l_{p}(\w, \bb)\!=\!\E_{\x,\y}(l_p(\w, \bb|\x,\y))\!=\!\E_{\x,\y,\e}(l_p(\w, \bb|\x,\y,\e)))$.
\end{itemize}
\end{defn}
In an ideal world, one would minimize  the ilf to obtain the estimation on $\w$ and $\bb$. The empirical version of the ilf is the elf, which is the objective function without any regularization. A NIRT minimizes the pelf, the expectation of which over the distribution of  noise is the  nm-pelf and is approximately equal to the elf with a penalty term to mitigate over-fitting in expectation (or as $k\rightarrow\infty$) as shown in Sections \ref{sec:regularization} and \ref{sec:sensitivity}.  
We also  establish a desirable behavior of the minimizer of pelf in an asymptotic sense as $n\rightarrow \infty$ and $k\rightarrow\infty$,  that is, it is consistent for the minimizer of the ilf (Theorem \ref{thm:arg}).  To that end, we  first establish the almost sure  (a.s.) convergence of the pelf to the ilf (Corollary \ref{lem:as4}) through the a.s convergence of the pelf to the nm-pelf (Lemma \ref{lem:as1}), from the nm-pelf to the fm-pelf (Lemma \ref{lem:as2}), and from the fm-pelf to the ilf (Corollary \ref{lem:as3}).  The relationships among the different loss functions and the main theoretical results are depicted in Figure \ref{fig:loss}. 

The lemmas and theorems presented below are based on the universal approximation theorem \citep{UA91}, which states $f$  is an universal approximator for the true underlying relation function between $\mathbf{Y}$ and $\mathbf{X}$  under mild regularity conditions. In addition, the establishment of the theoretical properties of whiteout also requires the Lipschitz continuity on the loss functions and compactness of the weight space.

\begin{lem}[\textbf{Almost sure convergence of pelf to nm-pelf}] \label{lem:as1}
In a NN of one hidden layer with bounded hidden nodes,
$|\inf\limits_{\w,\bb}l_{p}(\w, \bb|\x,\y,\e)-\inf\limits_{\w,\bb}l_{p}(\w, \bb|\x,\y)|<\delta$
as  $k\rightarrow\infty$ for any $\delta>0$ with probability 1.
\end{lem}
The proof is provided in Appendix \ref{app:as1}.   In a similar fashion as in  Lemma  \ref{lem:as1}, we also obtain Lemma   \ref{lem:as2}, the proof of which is provided in Appendix \ref{app:as2}.

\begin{lem}[\textbf{Almost sure convergence of nm-pelf to fm-pelf}]\label{lem:as2}
 In a NN of one hidden layer, where the hidden nodes are uniformly bounded, $|\inf\limits_{\w, \bb}l_{p}(\w, \bb|\x, \y)-\inf\limits_{\w, \bb}l_{p}(\w, \bb)|\!<\!\delta$ as $n\!\rightarrow\!\infty$  for any $\delta\!>\!0$ with probability 1.
\end{lem}
\noindent Given Lemmas \ref{lem:as1} and \ref{lem:as2}, together with the triangle inequality
\begin{align*}
&|\inf\limits_{\w, \bb}l_p(\w, \bb|\x,\y,\e)-\inf\limits_{\w, \bb}l_{p}(\w, \bb)|\\
\leq & |\inf\limits_{\w, \bb}l_p(\w, \bb|\x,\y,\e)-\inf\limits_{\w, \bb}l_p(\w, \bb|\x,\y)| +\\
& |\inf\limits_{\w, \bb}l_{p}(\w, \bb|\x,\y)-\inf\limits_{\w, \bb}l_{p}(\w, \bb)|,
 \end{align*}
 we can easily establish the a.s. convergence of the pelf to the fm-pelf (Corollary \ref{lem:as3}).
\begin{cor}\label{lem:as3}
\textbf{Almost sure convergence of pelf to fm-pelf:}
$|\inf\limits_{\w, \bb}l_p(\w, \bb|\x,\y,\e)- \inf\limits_{\w, \bb}l_p(\w, \bb)|<\delta$  as  $k\rightarrow\infty, n\rightarrow\infty$
for any $\delta>0$ with probability 1.
\end{cor}
With the results from the Lemmas \ref{lem:as1} and  \ref{lem:as2} and Corollary \ref{lem:as3}, we are now ready to establish the a.s. convergence of the pelf to the ilf (the proof is given in Appendix \ref{app:lem4}).
\begin{thm}\label{lem:as4}
\textbf{Almost sure convergence of pelf to ilf: } Let $\sigma_{\mbox{max}}(n)$  be the maximum noise variance among all injected noises.  If $\sigma_{max}(n)\rightarrow0 \mbox{ as }n\rightarrow\infty $, then $|\inf\limits_{\w, \bb}l_p(\w, \bb|\x,\y,\e)- \inf\limits_{\w, \bb}l(\w, \bb)|<\delta$ as $k\rightarrow\infty, n\rightarrow\infty$ for any $\delta>0$ with probability 1.
\end{thm}
Theorem  \ref{lem:as4} sets the foundation for Theorem \ref{thm:arg}, which is the main result that establishes whiteout as a reliable approach for learning a NN model under some regularity conditions.
\begin{thm}\label{thm:arg}
\textbf{Consistency of the minimizer of pelf to the minimizer of ilf: }
Let  $\hat{\w}_{p}^{r,n}$ and  $\w^0$ denote the optimal weight vectors from  minimizing the pelf $l_p(\w,\bb|\x,\y,\e)$, and the ilf $l_p(\w,\bb|\x,\y)$ respectively. Let  $r$ be the reciprocal of step length in an iterative weight updating  algorithm (e.g. $r$ is the reciprocal of the learning rate in the BP algorithm) and $r\rightarrow\infty$ (i.e., infinite noises are generated and injected during the weight training). Let $\W$ be the weight space, assumed  to be compact. Define $\hat{\W}^0=\{\w^0\in \W|l_p(\w^0,\bb)\leq l(\w,\bb) \mbox{ for all } \w \in \W\}$ that consists the minimizers of the ilf and is a non-empty subset of $\W$. Define the distance of $\w$ from $ \hat{\W}^0$ as $d(\w, \hat{\W}^0)=\min\limits_{\w^0\in \hat{\W}^0}||\w-\w^0||$ for any $\w\in \W$.  If
$\Pr\left(\!\sup\limits_{f\in \mathbf{F}_{n}}\big|l(\w, \bb)-l_{p}(\w, \bb|\x,\y,\e) \big|>t\!\right)\!\!\rightarrow\! 0$ as $r\!\rightarrow\!\infty$, $n\rightarrow\infty$, then
$\Pr\big(\lim\limits_{n\rightarrow\infty}\big(\limsup\limits_{r\rightarrow\infty}d( \hat{\w}_p^{r,n}, \hat{\W}^0)\big)=0\big)=1$
\end{thm}
The proof of Theorem \ref{thm:arg} is given in Appendix \ref{app:arg}. Note that Theorem \ref{thm:arg}  focuses on $\w$ only as the computation and estimation of $\bb$ are not affected by the NI. The condition that $r\rightarrow\infty$ is a stronger requirement than $k\rightarrow\infty$; as $r\rightarrow\infty$, we have $k\rightarrow\infty$, but not the other way around.
\begin{rem}
The proofs for Lemmas \ref{lem:as1} and  \ref{lem:as2},  Corollary \ref{lem:as3}, Theorem \ref{lem:as4}, and Theorem   \ref{thm:arg}  do not require the injected noises  to follow Gaussian distributions. Therefore, the a.s. convergence conclusions among the loss functions and the parameter consistency results should hold for all the NIRTs in addition to whiteout under the same regularity conditions.
\end{rem}

We have examined the asymptotic properties of the loss functions and their minimizers above.  When implementing whiteout in practice,  one minimizes the pelf with a finite number of epochs $k$ for a given training set of size $n$. It is important to  examine the fluctuation of pelf around  its expectation ($k\rightarrow\infty$) and it tail bound to ensure it is trainable.
\begin{cor}[\textbf{Tail bound on pelf}]\label{cor:tail}
 Assume output $\y$ is bounded; the loss function is uniformly bounded, and the activation functions employed by a NN are  Lipschitz continuous, 
then there exists a Lipschitz constant $B/\sqrt{kn}>0$, such that  $l_p(\w, \bb|\x,\y,\e): \mathcal{R}^{k}\rightarrow\mathcal{R}$, as a function of $\e_{k\times 1}$ where $\e$ are injected Gaussian whiteout noise, is $B/\sqrt{kn}$-Lipschitz with respect to the Euclidean norm, for any $ \delta>0$,
\begin{align}\label{eqn:tail}
&\Pr(\big| l_p(\w, \bb|\x,\y,\e)-\E_\e\left(l_p(\w, \bb|\x,\y,\e)\right)\big| > \delta)\notag\\
\leq&\; 2\exp\left(- kn\delta^{2}/(2B^2)\right).
\end{align}
\end{cor}
The proof of  Corollary \ref{cor:tail} is provided in Appendix \ref{app:tail}. Eqn (\ref{eqn:tail}) suggests that the fluctuation of pelf around its expectation is controlled in the sense that the distribution on the difference between pelf and its expectation nm-pelf has tails that decay to zero exponentially fast in $k$, providing assurance on the plausibility of minimizing the pelf for practical applications. 

In the establishment of the theoretical properties of whiteout, we have referred to several papers from the 1990’s: \citet{Grandvalet1997,Holmstrom1992} and \citet{Lugosi95}. Among the three, \citet{Lugosi95} does not have a NI component, but we borrow some of its framework to prove the consistency of the loss function minimizer with NI. \citet{Grandvalet1997} uses Gaussian NI, but its Gaussian noise is assumed to have a constant variance and the NI occurs only in the input layer, while the variance of whiteout noise is a function of the parameter estimates and changes during iteration (aka adaptive) and whiteout can inject noises into both input and hidden layers. In addition, whiteout noise can bring in sparsity regularization such as the $l_1$ or $l_{\gamma}$  for $0<\gamma<2$, while the constant-variance Gaussian noise can only impose the $l_2$ regularization. Finally, we borrow some of the framework in \citet{Holmstrom1992} to prove the tail bound property of the pelf. Again, \citet{Holmstrom1992} focuses on NI in the input layers only; in addition, it does not prove that the introduction of additive noise to the training vectors always improves network generalization. 

\section{Simulated Experiments}\label{sec:simulation}
In this section, we apply whiteout in simulated NN data to compare its prediction performance and regularization effects with dropout,  shakeout,  and  without regularization (referred to as ``no-reg'' hereafter). We choose dropout and shakeout to compare with whiteout since the former is the most widely used NIRT in practice, and the latter offers sparsity regularization ($l_1$). 

We focus on examining relatively large NN models with small training data, where the overfitting issue is typical. Specifically, we were interested in predicting a 5-category outcome and examined the following NN structures:  50-10-5, 50-15-5, 50-20-5, 50-15-10-5, and 70-10-5, where the first  number in each structure represents the number of input nodes, and the last number represents the number of output nodes, and the middle number(s) present the number(s) of hidden nodes in one or two hidden layers. The activation function  between the input and hidden layers was  sigmoid  and that between the hidden and the output layers was softmax. In the NNs with 50 input nodes, the weights in the NN were simulated from N$(0, 1)$ and the set that led to balanced outcomes  among the 5 categories was employed. In NN-70-10-5, there were 20 redundant nodes in the sense that the 200 weights associated with them were exactly 0. 
The input node values in the training set (100 cases) and testing data (100,000 cases) were drawn from $N(0,1)$ and the output nodes were calculated through the true NN, and the majority rule was applied to impute the 5-category outcome. We simulated 50 repetitions. When training the NNs, we set the learning rate at 0.2, the momentum at 0.5, and the number of epochs at 200,000. Under these settings, the training loss in no-Reg was on the order of $o(10^{-4})$ in all  the examined NN structures. The NIRTs were applied to the input nodes thus regularizing the weights between the input and hidden layers. 

We used a 4-fold CV to select tuning parameters in each NIRT. The tuning parameter $\tau$ in dropout is the probability of dropping an input node in this setting. For shakeout, one tuning parameter has the same interpretation as $\tau$, and the other  $c>0$ controls the relative weighting on $l_1$ and $l_2$ regularization.  We set $c=0.5$ (to yield a regularization effect of $l_1+l_2$), and applied the CV to select $\tau$. For whiteout, both additive and multiplicative, we first set $\gamma=1$ and used CV to select  $\sigma^2=\lambda$ to yield the $l_{2-\gamma}+l_2$ regularization. Once  $\sigma^2$ and $\lambda$ were  chosen, the CV was applied again to select $\gamma\in(0,2)$.  
Since the chosen $\gamma$ by CV in the additive whiteout noise case was around 1 without significantly affecting the prediction accuracy, to save computational time, we set $\gamma$ at 1 in the  multiplicative case and only tuned $\sigma^2=\lambda$. The final $\tau$ for dropout ranged from 0.05 to 0.07 across the repetitions and NN structures; $\tau$ for shakeout ranged from 0.4 to 0.6; $\gamma$ in the additive whiteout ranged from 0.8 to 1.0 and  $\sigma^2$ ranged from 0.4 to 1.2; $\sigma^2$ in the multiplicative whiteout ranged from 0.3 to 0.8 ($\gamma$ was fixed at 1).

The prediction accuracy  in the testing set in each examined NN structure was summarized over 50 repetitions and is presented in Table \ref{tab:simulation}.  First, all NIRTs improved the prediction accuracy compared to no regularization. Second, whiteout outperformed dropout and shakeout with the highest accuracy and the smallest SD (i.e., more stable across the repetitions).  Third, the additive whiteout seemed to deliver better performance overall. The only case where the multiplicative whiteout was better is NN-50-15-10-5, but its accuracy was very similar to the additive whiteout with $l_{2-\gamma}+l_2$. Furthermore, additive with the $l_{2-\gamma}+l_2$ regularization  was slightly better than the additive with the $l_1+l_2$ regularization. Fourth, in NN-70-10-5 where some weights were exactly 0, additive whiteout noise led to significantly better prediction than the other NIRTs.
\begin{table}[!htp]
\centering
\resizebox{\linewidth}{!}{
\begin{tabular}{@{}c@{\hskip3pt} c@{\hskip3pt}c@{\hskip3pt}c@{\hskip3pt}c@{\hskip3pt}c@{\hskip3pt}c@{}}
\hline
&\multicolumn{6}{c}{Prediction Accuracy: Average (SD) (\%) }\\
\cline{2-7}
                    &              &                &                 & multi. &add. &add.\\
                   & no-reg & dropout &shakeout & whiteout & whiteout & whiteout \\
 structure  &&$l_2$&$l_1+l_2$&$l_1+l_2$&$l_1+l_2$&$l_{2-\gamma}+l_2$\\
\hline
50-10-5    & 39.21 &42.71&41.89&\cellcolor{gray!25} 44.47&\cellcolor{gray!25} \textbf{44.55}& \cellcolor{gray!25} {44.51}  \\
                   & (1.81)&(0.66)&(1.83)&\underline{(0.30)}&(0.42)&(0.62)\\
50-15-5    &39.25 &  41.17 &44.19&44.74& \cellcolor{gray!25} 45.86& \cellcolor{gray!25} \textbf{46.09} \\
	               &(1.17)&(1.18)&(1.20)&(1.13)&(1.16)&\underline{(0.99)}\\
50-20-5    &39.69  &42.87&41.61&46.96&\cellcolor{gray!25} 47.37& \cellcolor{gray!25} \textbf{47.77}  \\
		          &(1.86) &(0.55)&(0.85)&(0.57)&\underline{(0.43)}&(0.48)\\
50-15-10-5& 36.03 & 39.80&39.25&\cellcolor{gray!25}\textbf{44.51}&42.47& \cellcolor{gray!25} 44.48\\
                     &(1.69) &(1.31)&(1.68)&\underline{(0.67)}&(0.96)&\underline{(0.67)}\\
70-10-5    &33.64 &40.53 &39.09&39.06&\cellcolor{gray!25} \textbf{44.98}&\cellcolor{gray!25} \textbf{44.98}\\
		           &(1.55)&(0.80)&(1.92)&(0.85)&\underline{(0.40)}&\underline{(0.40)}\\
\hline
\end{tabular}}
\caption{Averaged prediction accuracy in the testing data over 50 repetitions (\textbf{bold} represents the best accuracy in each NN, and the shaded cells present similar accuracy to the best accuracy)}\label{tab:simulation}
\end{table}

\begin{figure}[!htb]\begin{center}\includegraphics[ height=0.7\linewidth]{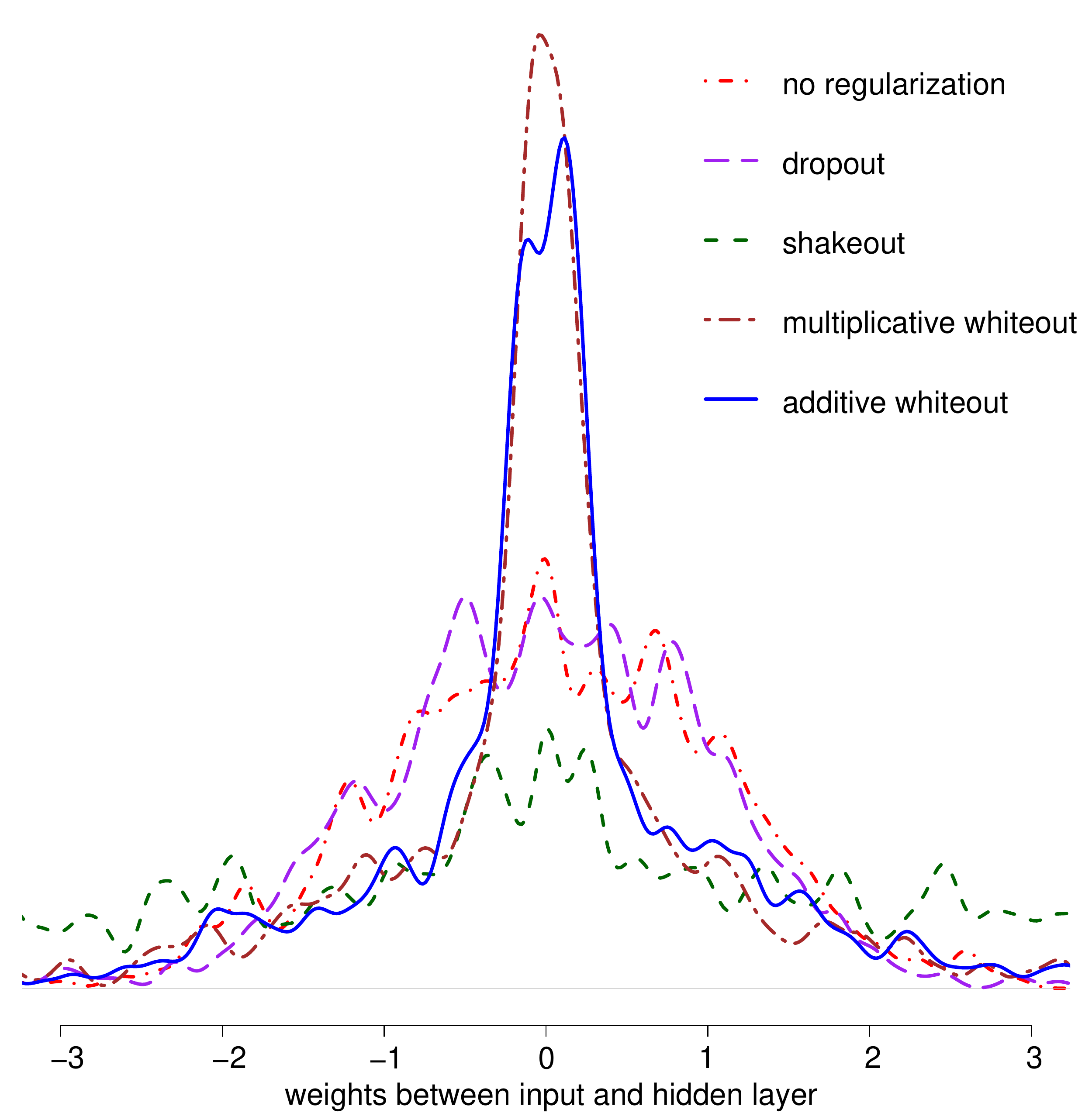}
\caption{Empirical distributions of the learned weights connecting the input and hidden layers  in NN-70-10-5 trained with simulated data in one repetition}\label{fig:sim1}
\end{center}\end{figure}

The distributions of the learned  weights between the input and hidden layers in the NN-70-10-5 structure are depicted in Figure \ref{fig:sim1}. The plot suggests that the whiteout (additive and multiplicative) were effective in introducing sparsity into the weight estimation, especially in this case where there are 20 ``redundant'' input whose weights were exactly 0's.

\section{Real-life Data Experiments}\label{sec:experiment}
In this section, we apply whiteout to four real-life data sets: the MNIST data, the CIFAR-10 data, the Lee Silverman Voice Treatment (LSVT)  voice rehabilitation data, and  the LIBRAS hand movement data. 
The LSVT and LIBRAS data sets are relatively small in size and are used to examine whether and how much the more flexible sparsity regularization in whiteout helps  in improving prediction accuracy compared to shakeout and dropout.  The MNIST and CIFAR-10 data are large and are used to benchmark the performance of whiteout against  dropout,  shakeout,  and no-reg with the same NN structure.  

The detailed results  are listed in Sections \ref{sec:lsvt} to \ref{sec:mnist} for each experiment. The common findings across the 4 applications are summarized as follows. First, all 3 examined NIRTs (dropout, shakeout, and whiteout) yielded higher prediction accuracy than no regularization. Second, whiteout outperformed shakeout and dropout when the training size was small (LSVT and LIBRAS), and delivered comparable performance  shakeout and dropout when the training size was large (MNIST and CIFAR-10). Third, for image classifications in MNIST and CIFAR-10 where  the state-of-art NNs were employed, the advantage of whiteout and shakeout over dropout was not obvious. In the trace plots of the training loss across epochs, there was more fluctuation in the loss curves for shakeout than whiteout, both of which had more fluctuation than dropout given the noise injected depended on the weight updates themselves in the former two.

\subsection{LSVT voice rehabilitation data}\label{sec:lsvt}
The LSVT data set contains 309 dysphonia measures  attributes from 126 samples of 14 participants with Parkinson's disease (PD) who underwent LSTV for speech disorders. 
After LSTV, the  phonation was assessed by LSVT expert clinicians and  labeled   as ``acceptable''  or  ``unacceptable''.  The goal was to predict the phonation outcome. The LOGO/SVM algorithm employed in the original paper yielded a prediction validation accuracy around 85\% to 90\%, and with SD around $8\%\sim 10\%$  (Figure 2 in \citet{lsvt}).

The NN model we applied to the LSVT data  contained  two fully connected hidden-layers with the sigmoid activation function.  The first hidden layer contained 9 nodes, and the second had 6 nodes. The 309 attributes were standardized before being fed  to the input layer of the NN model. In terms of the tuning parameters, we examined several scenarios for $\sigma$ in whiteout (listed in Table \ref{tab:exp1result}).  The larger $\sigma$ was, the more dispersed the noises were. We set  $\lambda=\sigma^2$ and $\gamma=1$ and to yield the EN type of regularization in whiteout. According \citet{Kang2016}, Gaussian NI with a constant variance leads to the same $l_2$ regularization as dropout if the variance in the former is set at $\tau/(1-\tau)$. To make the $l_2$ regularization in dropout to comparable between dropout and whiteout, we set $\tau$ in dropout at $\tau=2\sigma^2/(1+2\sigma^2)$  (2 because both $\sigma^2$ and $\lambda$, which were set to be equal, contributed to the $l_2$ regularization in whiteout). Similarly, to make the regularization effect comparable between whiteout and shakeout, we set $c=0.5$ and calculated $\tau$ as $2\sigma^2/(1+2\sigma^2)$. Similar to the original paper, we run 100 repetitions  with a 10-fold CV (about 13 samples in each validation set) and calculated the mean accuracy across the repetitions. The initials weights were randomly sampled from  $N(0,1)$. We run 100,000 epochs with a learning rate of 0.15 and a momentum of 0.15  in each learning algorithm. 

The results are given in Table \ref{tab:exp1result}.  All NIRTs  led to better prediction accuracy than the regular BP without regularization, regardless of the noise level. Whiteout had the best performance with the highest prediction accuracy (bolded in the table)  among all NI methods in all the examined noise variance scenarios, which was comparable to the  LOGO+SVM  algorithm employed by the original paper. When $\sigma=0.4$ in whiteout, according to $\tau\!=\!4/29\!$ in dropout and $\tau\!=\!8/33$, the accuracy was the highest in each NIRT. In addition, smaller SDs of the accuracy rates were achieved in either whiteout or shakeout perhaps due to the additional $l_1$ regularization compared to dropout (except for $\sigma\!=\!0.7$). In addition, it is interesting that the multiplicative whiteout noise delivered better performance than its additive counterpart if $\sigma$ (the same between the two), was small; but worsen as $\sigma$ got larger.
\begin{table}[!htp]
\centering
\resizebox{\linewidth}{!}{
\begin{tabular}{c@{\hskip3pt}c@{\hskip3pt}c@{\hskip3pt}|@{\hskip2pt}c@{\hskip3pt}c@{\hskip3pt}c@{\hskip3pt}c@{\hskip3pt}c@{}}\hline
\multicolumn{3}{c|}{tuning parameters}& &  & & multi. & add. \\
\cline{1-3}
white- & shake-& drop-  &  no & drop & shake & white& white\\
out ($\sigma$) & out ($\tau$) & out ($\tau$)& -reg  & -out &  -out  &   -out &   -out   \\
 \hline
0.2 & 1/26 & 2/27 & &79.34 & 85.10 &  \textbf{85.35} & 83.37\\
 &  & & &(8.84) &   (8.75) & (8.74) &(9.29) \\
0.3 & 9/109 & 9/59 & &83.42 & 84.61  &\textbf{85.26} & 81.55\\
 &  & & & (7.82) &  (9.09) &(8.46) &(9.58) \\
0.4 & 4/29 & 8/33 & 73.93 &   \underline{84.17} &  \underline{84.66} & \underline{\textbf{87.27}} & 82.22\\
 &  & & (12.13) & (10.37) & (8.98) & (9.84) &(9.67) \\
0.5 &1/5 & 1/3 & &82.95  & 83.88 & \textbf{85.99} &83.56\\
 &  & & &  (9.86) & (8.46) &  (9.37) & (10.22)\\
0.6 & 9/34 & 18/43& & 81.19 & 83.92  &84.22 & \textbf{86.35}\\
 &  & & &  (10.16) &  (8.75) & (6.05)& (9.28) \\
0.7 & 49/149 & 49/99 & & 80.76  & 81.25 & 84.62 & \textbf{85.31} \\
 &  & & &  (10.24) & (11.86) & (11.34) &(7.95)\\\hline
\end{tabular}}
\caption{Mean (SD) prediction accuracy rates (\%)  in the validation data  over 100 repetitions in  the 10-fold CV  in the LSVT data}\label{tab:exp1result}
\end{table}

We also examined the empirical distribution of the weight estimates from the input layer (due to space limitation, we present the plot in Figure \ref{fig:pdfLSVT} of Appendix \ref{sec:density}). With the additional $l_1$ regularization term in whiteout and shakeout, more estimated weights were around 0 compared to dropout ($l_2$ regularization only) and no regularization.


\subsection{LIBRAS movement data}\label{sec:libras}
LIBRAS, the acronym for Portuguese  ``L\'{i}ngua BRAsileira de Sinai'', is the official Brazilian sign language. The LIBRAS data set contains 15 commonest hand movement categories that is represented as a two-dimensional curve, and contains 60 samples with 90 attributes. The unsupervised Self Organizing Map (SOM) and (fuzzy) Learning Vector Quantization (LVQ) algorithms employed in the original paper yielded the maximum accuracy rates (\%) of 55.56, 57.78, and 88.79 in three test sets.   \citet{moveurl} applied the $k$ nearest neighborhood (nn) algorithm, and the highest average accuracy rate (\%) achieved was 39.2\% (SD = 12.7).

The NN we employed contained two fully connected hidden layers NN (with 20 and 10 nodes, respectively) with the sigmoid activation functions connecting the input and the first hidden layer, and the softmax function connecting the second hidden layer and the output layer. We varied the size of the training set from 60 to 240  and set side 120 as the testing set.  The input data were standardized before fed to the NN. In whiteout, we set $\gamma=1$ and applied the 4-fold CV to select $\sigma^2=\lambda$; in shakeout, we set $c=0.5$ and used the 4-fold CV to select $\tau$; in dropout, $\tau$ was chosen via the 4-fold CV. The initial weights were randomly sampled from N$(0,1)$. The number of epochs  was 200,000 with a learning rate of 0.2 and a momentum of 0.2 in each BP algorithm with or without NI. The prediction accuracy in the test set was summarized over 50 repetitions and is presented in Table \ref{tab:exp2result}. 
\begin{table}[!htp]
\resizebox{\linewidth}{!}{\centering
\begin{tabular}{@{}c@{\hskip2pt}|@{\hskip2pt}c@{\hskip4pt}c@{\hskip4pt}c@{\hskip4pt}c@{\hskip4pt}c@{}}\hline
training & regular BP & dropout & shakeout & whiteout \\
 size ($n$) & &&\\
\hline
60 &  37.46 (4.22) &42.59 (3.59)& 42.86 (3.47) & \textbf{48.54} (3.41) \\
120 & 51.57 (3.58) & 55.74 (3.93) &55.20 (4.06) & \textbf{61.11} (3.24)\\
180 & 53.55 (5.26) & 55.83 (4.22) & 56.27 (4.17) & \textbf{61.56} (3.08) \\
240 & 62.15 (5.73)&66.35 (3.60) & 66.85 (3.56) & \textbf{69.04} (2.08) \\
\hline
\end{tabular}}
\caption{Mean (SD) prediction accuracy (\%)  in the testing data with different training sizes  over 50 repetitions  in the LIBRAS data}\label{tab:exp2result}
\end{table}

Whiteout had the best performance --  the highest accuracy rates and the smallest SDs -- among the three NIRTs across all the examined  training size scenarios.   In this  application, shakeout improved very little over the dropout procedure, which was also implied by the similarity of the final weight estimate distributions from two procedures (Figure \ref{fig:pdfmove} in Appendix \ref{sec:density}). By contrast, more weights were estimated around 0 in whiteout. When the training size was 180 (the same as the original paper though not necessarily the same set), whiteout yielded a 61.56\% accuracy rate, better than the fuzzy LVQ algorithm \citep{move} and the $k$-nn method \citep{moveurl}.

\subsection{MNIST and CIFAR-10}\label{sec:mnist}
The  MNIST  and CIFAR-10 data are classic data for testing and benchmarking image classifiers. The MNIST data comprise of hand written digits from 0 to 9. The CIFAR-10 consists of 60,000 $32\times32$ color images in 10 classes 
with 6,000 images per class. The goal of these two experiments is  not to design new NNs  to beat the best classifiers out there, but rather to compare the regularization effect of whiteout, dropout, and shakeout when the same NN is used.

We employed the same NN structure (Tables \ref{tab:mnist1} and  \ref{tab:cifar1} in Appendix \ref{sec:NNstructure}) as employed in the shakeout paper \citep{Kang2016}. The 3 NIRTs were applied to hidden nodes in the fully connected layers  (layers 3 and 4), and a 4-fold CV  was applied to select the tuning parameter in each NIRT with 50 repetitions. We run 200 epochs with a learning rate 0.0005. 
The prediction accuracy in the  testing data was  summarized over the 50 repetitions  and is shown in Table \ref{tab:mnist2} for the MNIST experiment and Table \ref{tab:cifar2} for the CIFAR-10 experiment. In summary,  all NIRTs yielded better predictions than no regularization, but there was minimal difference in both the accuracy rate and  SD among the 3 NIRTs as the tuning parameters selected by the CV led to only a small amount of noise injected in each NIRT, especially when $n$ was large.
\begin{table}[!htp]
\centering\resizebox{0.9\linewidth}{!}{
\begin{tabular}{@{}r|@{\hskip2pt}c@{\hskip5pt}c@{\hskip5pt}c@{\hskip5pt}c@{}}
\hline
$n$ & no-reg & dropout & shakeout & whiteout \\\hline
500 & 9.77 (0.21) & 5.95 (0.25) & 6.06 (0.23) & 6.02 (0.27) \\
1,000 & 6.53 (0.19) & 4.19 (0.23) &4.26 (0.26)& 4.21 (0.24)\\
3,000 & 3.01 (0.15) &  2.01 (0.18) & 2.07 (0.20) & 2.04 (0.23) \\
8,000 & 1.79 (0.08) & 1.12 (0.11) & 1.05 (0.12) &  1.10 (0.11) \\
20,000 & 1.01 (0.04) & 0.92 (0.05) & 0.95 (0.08) & 0.93 (0.09)\\
50,000 & 0.81 (0.03) & 0.77 (0.03) & 0.76 (0.05) & 0.78 (0.08)\\
\hline
\end{tabular}}
\caption{Mean (SD) Misclassification rates (\%) with different training sizes in the MNIST data }\label{tab:mnist2}
\end{table}
\begin{table}[!htp]
\centering\resizebox{0.95\linewidth}{!}{
\begin{tabular}{@{}r@{\hskip2pt}|@{\hskip1pt}c@{\hskip5pt}c@{\hskip5pt}c@{\hskip5pt}c@{}}
\hline
$n$ & no-reg &  dropout  &  shakeout  & whiteout \\\hline
300 &70.30 (1.34)&65.52 (1.52) & 64.69 (1.76) & 64.58 (1.68) \\
700 &62.24 (0.97)& 57.12 (1.14) &56.83 (1.33) & 57.05 (1.37)\\
2,000 & 53.60 (0.69)& 48.48 (0.91)&47.64 (1.06) & 47.87 (1.12) \\
5,500 & 44.54 (0.56)& 39.75 (0.69) &39.88 (0.93) & 40.04 (0.77) \\
15,000 & 33.40 (0.41)& 29.26 (0.62) & 30.12 (0.61) & 30.09 (0.74) \\
40,000 & 24.78 (0.16)& 22.02 (0.22) & 22.15 (0.58) & 21.43 (0.39) \\
\hline
\end{tabular}}
\caption{Mean (SD) Misclassification rates (\%)  with different training sizes in the CIFAR-10 data}\label{tab:cifar2}
\end{table}

\section{Discussion}\label{sec:discussion}
We have examined the Gaussian NIRT for NNs through designing of whiteout, a family of NIRTs that injects adaptive Gaussian noises into input and hidden nodes in a NN during training. Whiteout has connections with the bridge, lasso, ridge, and EN  penalization in GLMs, and can also be extended to offer regularization similar to the adaptive lasso and group lasso.  
The versatility of the whiteout can be ascribed to the adaptive variance of the injected Gaussian noise, which is a function of weight parameters and contains three tuning parameters. It appears at first that tuning three parameters makes whiteout less attractive given the possibly high computational cost on parameter tuning, say via a grid search through CV. However, this should not be a concern as tuning parameters is embarrassingly parallel computationally. In addition,  tuning all three is often not necessary. As  stated in Section \ref{sec:tuningparameter}  and demonstrated in the experiments,  users may first decide on what type of  regularization effects  they would like to achieve when training a NN, which will help fix one or two tuning parameters at specific values and leave only two or one for tuning. For example, if we desire a bridge-type penalty,  we may set $\lambda=0$, leaving $\sigma^2$ and $\gamma$ for tuning;  if we desire a EN-type penalty,  then we can set $\gamma=1$, leaving $\sigma^2$ and $\lambda$ for tuning. If a user has difficulty to decide,

We have presented both additive and multiplicative whiteout noises. The experiment results do not suggest one is always superior to the other in terms of prediction performance. We have briefly mentioned in Section \ref{sec:multiplicative} that the multiplicative noise might have additional robustness effects for weakening the impact of the ``outlying'' nodes on the prediction. We plan to explore this issue  methodologically and empirically in the future to see if the conjecture holds.

Whiteout can also be applied to unsupervised learning  such as dimension reduction and pre-training of deep NNs. Appendix \ref{sec:unsupervised} illustrates the regularization effects of the whiteout in RBMs and auto-encoders, both of which are widely  used unsupervised learning methods.  In both cases, the expected perturbed loss functions with regard to the distribution of whiteout noises can be approximated as the original loss function plus a penalty term.


\bibliographystyle{IEEEtranN}
\input{arxiv4.bbl}

\appendices
\section{Proof of Lemma \ref{lem:Rw}}\label{app:Rw}
\begin{align*}
& \textstyle\sum_{i=1}^{n}\!\mbox{E}_{\e}(l(\w|\tilde{\x}_i,y_i))\\
= & \textstyle-d(\!\tau\!)^{-1}\!\sum_{i=1}^{n}\!(\mathbf{T}(y_i)\mbox{E}_{\e}\!(\tilde{\x}_i)\w\!-\!\mbox{E}_{\e}(\!A(\tilde{\x}_i\w)))\! -\!\mbox{log}(h(Y,\tau))\\
=& \textstyle-d(\tau)^{-1}\sum_{i=1}^{n}\left(\mathbf{T}(y_i)\x_i\w\!-\!\mbox{E}_{\e}(A(\tilde{x}_i\w))\right)\!-\!\mbox{log}(h(Y,\tau)\\
=& \textstyle \sum_{i=1}^{n}l(\w|\x_i,y_i)+R(\w)/d(\tau),
\end{align*}
where $R(\w)\triangleq\sum_{i=1}^{n}\!\mbox{E}_{\e}(A(\tilde{\x}_i\w))\!-\!A(\x_i\w)$. The expectation of a second-order Taylor expansion of $A(\tilde{\x}_i\w)$ around $\x_i\w$  with regard to the distribution of noise  leads to
$\mbox{E}_{\e}(A(\tilde{\x}_i\w))\approx A(\x_i\w)+A'(\x_i\w)\mbox{E}_{\e}(\tilde{\x}_i\w-\x_i\w)+ \frac{1}{2}A''(\x_i\w)\mbox{V}_{\e}(\tilde{\x}_i\w\!-\!\x_i\w)
\!\!=\!\!A(\x_i\w)\!+\!\frac{1}{2}A''(\x_i\w)\mbox{V}_{\e}(\tilde{\x}_i\w)$. Thus, $R(\w)\approx \frac{1}{2} \sum_{i=1}^{n}A''(\x_i\w)\mbox{V}_{\e}(\tilde{\x}_i\w)$.

\section{Proof of Theorem \ref{thm:sensitivity}}\label{app:sensitivity}
Denote the number of layers by $L$ in a NN. Denote the additive whiteout noise injected to the NN by $\e=\{e_{ijk}\}$, where $e^{(l)}_{ijk}\sim\mbox{N}(0,\sigma^2|w_{jk}^{(l)}|^{-\gamma}+\lambda)$ for $i=1,\ldots,n; j=1,\ldots,m^{(l)}$ and $k=1,\ldots,m^{(l+1)}$ for $l=1,\ldots,L-1$ (we also use $p$ to denote the number of nodes in the input layer; that is, $p=m^{(1)}$). Given $\x_i$, the predicted outputs of dimension $q$ from the NN learned with the whiteout NI is
\begin{align*}
\hat{\y}_i&=\f\left(\x_i,\e_i\big|\w,\!\bb\right)
\end{align*}
where $\f=\f^{(L-1)}\circ \f^{(L-2)}\circ\cdots\circ \f^{(1)}=\f^{(L-1):1}$ is the vector of the compound functions over the series of active functions connecting layers 1 to $L$. The NN parameters $\w$ and $\mathbf{b}$ are estimated by minimizing the $l_2$ loss between the observed and predicted outcomes $\sum_{i=1}^{n}\!|\y_i-\hat{\y}_i|^2$ ($|\cdot|$ is the Euclidean norm).

Suppose there is a small external perturbation in the input nodes, denoted by $\mathbf{d}_{p\times1}$, that has a mean of $\mathbf{0}$ and covariance $\varpi^2 I_{p\times p}$. The predicted outcomes from  the learned NN with whiteout NI given the set of externally perturbed inputs is
$$ \hat{\hat{y}}_i\!=\f\left(\x_i+\mathbf{d}_i, \e_i|\w,\!\bb\right).$$
\noindent The change in the predicted output with vs. without the externally perturbation $\mathbf{d}$ is thus given by
\begin{align*}
\boldsymbol{\Delta}_i=\hat{\hat{\y}}_i-\hat{\y}_i.
\end{align*}
Denote the dimension of $\hat{\y}_i$ by $q$. Each of the $q$ elements in $\boldsymbol{\Delta}_i$ can be approximated (Eqn \ref{eqn:deltai}) through the first-order Taylor expansion  at $\x_i$ w.r.t. $\mathbf{n}_{q',i}=\!(\mathbf{d}_i+\e_{i1}^{(1)}\!,\ldots,\mathbf{d}_i+\e_{ip}^{(1)},\e_i^{(2)},\ldots,\e_i^{(L-2)},\e_{q',i}^{(L-1)})$ for $q'=1,\ldots,q$ ($\mathbf{n}_{q',i}$ is a column vector of length $pm^{(1)}+\sum_{l=1}^{L-2}m^{(l)}m^{(l+1)}+m^{(L-1)}q$)
\begin{align}\label{eqn:deltai}
&\boldsymbol{\Delta}_i\approx {\Psi}_i(\w,\bb)\cdot\mathbf{n}_i, \mbox{ where }
\end{align}
$${\Psi}_i(\w,\bb)^T =\begin{pmatrix}
\!\frac{\partial \f_1^{(L-1):1}}{\partial{f}_{1,1}^{(1)}} \frac{\partial{f}_{1,1}^{(1)}}{\partial{\x}_i} &
\ldots &
\frac{\partial \f_q^{(L-1):1}}{\partial{f}_{q,1}^{(1)}} \frac{\partial{f}_{q,1}^{(1)}}{\partial{\x}_i} \\

\vdots & \vdots & \vdots\\

\frac{\partial \f_1^{(L-1):1}}{\partial{f}_{1,m^{(2)}}^{(1)}} \frac{\partial{f}_{1,m^{(2)}}^{(1)}}{\partial{\x}_i} &
\ldots &
\frac{\partial \f_q^{(L-1):1}}{\partial{f}_{q,m^{(2)}}^{(1)}} \frac{\partial{f}_{q,m^{(2)}}^{(1)}}{\partial{\x}_i}\\

\frac{\partial\f_1^{(L-1):2}}{\partial{f}_{1,1}^{(2)}} \frac{\partial{f}_{1,1}^{(2)}}{\partial{\h}^{(1)}_i} & \ldots &
\frac{\partial\f_q^{(L-1):2}}{\partial{f}_{q,1}^{(1)}}
\frac{\partial{f}_{q,1}^{(2)}}{\partial{\h}^{(1)}_i}\\

\vdots & \vdots & \vdots\\

\frac{\partial \f_1^{(L-1):2}}{\partial {f}_{1,m^{(3)}}^{(2)}} \frac{\partial{f}_{1,m^{(3)}}^{(2)}}{\partial{\h}^{(1)}_i}& \ldots & \frac{\partial f_q^{(L-1):2}}{\partial {f}_{q,m^{(3)}}^{(2)}}
\frac{\partial{f}_{q,m^{(3)}}^{(2)}}{\partial{\h}^{(1)}_i}\\

\vdots & \vdots & \vdots\\

\frac{\partial f_1^{(L-1)}}{\partial{\h}^{(L-1)}_i} & \ldots & \frac{\partial f_q^{(L-1)}}{\partial{\h}^{(L-1)}_i}\\
\vdots & \vdots & \vdots\\
\frac{\partial f_1^{(L-1)}}{\partial{\h}^{(L-1)}_i}& \ldots & \frac{\partial f_q^{(L-1)}}{\partial{\h}^{(L-1)}_i}\\,
\end{pmatrix}$$
$\frac{\partial {f}_j^{(1)}}{\partial {\x}_i}\!=\!\left(\!\frac{\partial {f}_j^{(1)}}{\partial {x}_{i1}},\!\cdots,\!\frac{\partial {f}_j^{(1)}}{\partial {x}_{ip}}\!\right)^T$ for $j=1,\ldots,m^{(2)}$, ${\h}_i^{(l)}$ denotes the hidden nodes in the $l$-th layer, and $\frac{\partial {f}_j^{(l)}}{\partial {\h}^{(l)}_i} \!=\!\left(\!
\frac{\partial {f}_j^{(l)}}{\partial {h}^{(l)}_{i1}},\!\cdots,\!
\frac{\partial {f}_j^{(l)}}{\partial {h}^{(l)}_{i,m^{(l)}}}\!\right)^T\!\!\!$ for $j\!=\!1,\ldots,m^{(l)}$ and $l\!=\!2,\ldots,L-1$. For notation simplicity, we use ${\Psi}_i$ in place of ${\Psi}_i(\w,\bb)$, and the $q'$-th column of ${\Psi}_i$ by ${\Psi}_{q'i}$ in what follows.

We modify the sensitivity definition of a learned NN from \citet{NI92} to accommodate the training of the NN with the whiteout NI. The modified sensitivity $S(\w,\bb)$ is the summed ratio (over all cases $i=1,\ldots,n$) between the variance of $|\Delta_i|$ over the joint distribution of $\mathbf{n}_i$  and the variance of the total input perturbation into case $i$.
\begin{align}
&S(\w,\bb)=\!\sum_{i=1}^{n}\!\frac{\mbox{V}_{\mathbf{n_i}}(|\Delta_i|)}{\mbox{V}_{\mathbf{d}}(|\mathbf{d}|)}\notag\\
&\approx\sum_{i=1}^{n}
\frac{\sum_{q'=1}^q\E_{\mathbf{n}_{q',i}}({\Psi}_{q',i}\mathbf{n}_{q',i})^2}{\E_{\mathbf{d_i}}|\mathbf{d_i}|^2}\notag\\
&=p^{-1}\sum_{i=1}^{n}\!\sum_{q'=1}^q\!{\Psi}_{q',i}\!
\begin{pmatrix}
\!R\!+\!D_1\!\!&\!\!\! 0\!\!\\
\!0\!\!&\!\!\!D_{q',2}\!\!\\
\end{pmatrix}
\!{\Psi}_{q',i}^T\notag\\
&=\textstyle p^{-1}\sum_{i=1}^{n}\!\sum_{q'=1}^q\!{\Psi}_{q',i}D{\Psi}_{q',i}^T,\label{eqn:S}
\end{align}
where $R$ is a symmetric band matrix  that captures the correlation among $\mathbf{n}_{q',i}$ due to $\mathbf{d}_i$ and $R[i,i\!+\!p]\!=\!1$ for $i=$ \\$1,\ldots,p(m-1)$ and 0 otherwise in its upper triangle,
$D_1\!=\!\mbox{diag}\!\left(\!\sigma^2\varpi^{-2}\big|w_{jk}^{(1)}\big|^{-\gamma}\!\!\!+\!\lambda\varpi^{-2}\!+\!1\right)$ for $j\!=\!1,\ldots,p$ and \\$k= 1,\ldots,m^{(2)}$,
$D_{q',2}\!=\!\mbox{diag}\!\left(D^{(2)},\ldots,D^{(L-2)},D_{q'}^{(L-1)}\right)$  \\ with $D^{(l)}\!=\!\mbox{diag}\!\left(\!\sigma^2\varpi^{-2}\big|w_{jk}^{(l)}\big|^{-\gamma}\!\!\!+\!\lambda\varpi^{-2}\!\right)$ for $l=2,\ldots, L-2$ \\ and
$D_{q'}^{(L-1)}=\mbox{diag}\left(\sigma^2\varpi^{-2}\big|w_{jq'}^{(L-1)}\big|^{-\gamma}\!\!\!+\!\lambda\varpi^{-2}\right)$.

Following the framework in \citet{NI92},  we minimize the sum of the loss function and the sensitivity of the network, instead of the raw $l_2$ loss function $l(\w,\bb)=\sum_{i=1}^{n}\!|\y_i-\bar{\y}_i|^2$ alone ($\bar{\y}_i$ is the predicted outcome from minimizing $l(\w,\bb)$), which is prone to overfitting or instability in this context. That is, the objective function is
\begin{equation}\label{eqn:lossS}
\textstyle\sum_{i=1}^{n}\!|\y_i-\bar{\y}_i|^2+a S(\w,\bb),
\end{equation}
where $a$ is a tuning parameter. Plugging in $S(\w,\bb)$ from Eqn (\ref{eqn:S}), Eqn (\ref{eqn:lossS}) becomes
\begin{equation}\label{eqn:lossS1}
\textstyle\sum_{i=1}^{n}\!|\y_i-\bar{\y}_i|^2\!+a\sum_{i=1}^{n}\sum_{q'=1}^q{\Psi}_{q',i} D {\Psi}_{q',i}^T/p
\end{equation}
We now show that the objective function in Eqn (\ref{eqn:lossS1}) is approximately equivalent to $\E_{\e^*}(l_p(\w,\bb|\e^*,\x,\y))$, the expected loss function over the distribution of the sum of two sets of whiteout noise  $\e_{i}^*=\e_{i}^{a}\!+\!\e_{i}^{b}$ . Specifically,
\begin{align*}
\mbox{for }&\mbox{input nodes:}\\
&e_{ijk}^{a}\!\sim N(0,\sigma^{*2}\big|w_{jk}^{(1)}\big|^{-\gamma}+\lambda^{*})\mbox{ and } e_{ij}^{b}\!\sim N(0,ap^{-1});\\
\mbox{for } &\mbox{hidden nodes in layer $l=2,\ldots,L-1$}\\
&e_{ijk}^{a}\!\sim N(0,\sigma^{*2}\big|w_{jk}^{(l)}\big|^{-\gamma}+\lambda^{*})\mbox{ and } e_{ijk}^{b}= 0,
\end{align*}
where $\sigma^{*2}= a\sigma^2\varpi^{-2}/p$ and $\lambda^{*}=a\lambda\varpi^{-2}/p$. It can be show easily that $\mbox{V}(\e^*_i)=D$. In other words, Eqn (\ref{eqn:lossS1}) can be written in terms of $\e^*$
\begin{align*}
&\textstyle\sum_{i=1}^{n}|\y_i-\bar{\y}_i|^2+
\sum_{i=1}^{n}\sum_{q'=1}^q{\Psi}_{q',i}\mbox{V}(\e^*_i)\Psi^T_{q',i}\\\
= &\textstyle\sum_{i=1}^{n}|\y_i-\bar{\y}_i|^2+
\sum_{i=1}^{n}\sum_{q'=1}^q\mbox{V}(\Psi_{q',i}\e^*_i)\\
=&\textstyle\sum_{i=1}^{n}|\y_i-\bar{\y}_i|^2+
\sum_{i=1}^{n}\sum_{q'=1}^q\E(\Psi_{q',i}\e^*_i)^2\\
=&\;\textstyle\E_{\e}\big(\sum_{i=1}^{n}\sum_{q'=1}^q\left((y_{i,q'}-\bar{y}_{i,q'})-(\Psi_{q',i}^T\e^*_i)\right)^2\big).\\
\mbox{Per }&\mbox{the first-order Taylor expansion}\\
\approx&\;\textstyle \E_{\e^*}\!\big(\sum_{i=1}^{n}\sum_{q'=1}^q \big(y_{i,q'}-\hat{y}_{i,q'}-(\hat{y}_{i,q'}- \bar{y}_i)\big)^2 \big) \\
=&\;\textstyle\E_{\e^*}\big(\sum_{i=1}^{n}\sum_{q'=1}^q\big(y_{i,q'}-\hat{y}_{i,q'}\big)^2\big)\\
=&\;\E_{\e^*}(l_p(\w,\bb|\e^*,\x,\y)),
\end{align*}
implying minimizing the sum of the original loss function and the sensitivity of the NN is approximately equivalent to minimizing the whiteout perturbed loss function.

\section{Proof of Lemma \ref{lem:as1}}\label{app:as1}
This lemma is established in NNs with a  single hidden layer and a single output node.  Since
$|\inf\limits_{\w, \bb}l_{p}(\w, \bb|\x, \y,\e)-\inf\limits_{\w, \bb}l_{p}(\w, \bb|\x,\y)|\leq2 \sup\limits_{\w, \bb} |l_{p}(\w, \bb|\x,\y,\e)-l_{p}(\w, \bb|\x,\y)|$, we can establish the convergence of $\sup\limits_{\w, \bb} |l_{p}(\w, \bb|\x,\y,\e)-l_{p}(\w, \bb|\x,\y)|$, which would automatically imply the convergence of $|\inf\limits_{\w, \bb}l_{p}(\w, \bb|\x, \y,\e)-\inf\limits_{\w, \bb}l_{p}(\w, \bb|\x,\y)|$. By Lemma 2 in \citet{Lugosi95}, for any $t>0$,
\begin{align*}
&\Pr\left(\sup\limits_{\w, \bb} |l_{p}(\w, \bb|\x,\y,\e)-l_{p}(\w, \bb|\x,\y)| > t\right)\\
\leq& 4 \E (\mathbf{N}(t/16, \mathcal{L}(\e))\exp\left(-\frac{kt^{2}}{128B}\right),
\end{align*}
where $\mathcal{L}(\e)\!=\!\{(l_{p}(\w, \bb|\x,\y,\e_{1}),\ldots ,l_{p}(\w, \bb|\x,\y,\e_{k})); l_p\in \mathcal{L}\}\subseteq \mathcal{R}^{k}$ is the space of functions in $\mathcal{L}$ restricted to $\e_{1},\ldots, \e_{k}$, $B$ is the uniform bound on $\mathcal{L}(\e)$,  $\mathbf{N}(t/16, \mathcal{L}(\e))$ is the $L_{1}$ covering number of $\mathcal{L}(\e)$, defined as the cardinality of the smallest finite set in $R^{m(p+1)}$ ($p$ is the number  of the input nodes , and  $m$ is the  hidden nodes),  such that for every $a\!\in\!\mathcal{L}(\e)$ there is a point $a'\in \mathcal{R}^{m(p+1)}$ in this finite set such that $\frac{1}{m(p+1)}\big||a-a'|\big|_1\leq t$. By Theorem 1 in \citet{Lugosi95}, we assume without loss of generality that the inverse activation function of output variable $\y$ plus the summed noises injected into hidden nodes,  $(f^{(2)})^{-1}(\y)+\sum\e$,  is bounded upward (the inner products of weights and $\sum\e$ are practically unlikely to take on large values, if the BP algorithm is used for computation).   Define the NN model $\mathcal{F}$ as
\begin{align*}
\mathcal{F}=&\left\{\sum_{j=1}^{m}w_j^{(2)}f^{(1)}(\x \w_j^{(1)}+b^{(1)})+b^{(2)};\right.\\
& \left.\w_j^{(1)}\in \mathcal{R}^p, b^{(1)}, b^{(2)}\in \mathcal{R}, \sum_{j=1}^{m}|w_j^{(2)}|\leq \beta \right\}.
\end{align*}
Further assume that hidden nodes $f^{(1)}(\w_j,b_j)$ are uniformly bounded, then by Theorem 3 in \citet{Lugosi95},
\begin{align}\label{eqn:ineq1}
&\Pr\left(\sup\limits_{\mathcal{F}} |l_{p}(\w, \bb|\x,\y)-l_{p}(\w, \bb|\x,\y,\e)| > t\right)\notag\\
\leq&\; 8(512e\beta^3/t)^{2m}\exp\left(\frac{-knt^2}{2048\beta^4}\right),
\end{align}
which goes to 0 as $k\rightarrow\infty$, so does $\Pr\left(|\inf\limits_{\w, \bb}l_{p}(\w, \bb|\x, \y,\e)-\inf\limits_{\w, \bb}l_{p}(\w, \bb|\x, \y)| > t\right)$. By the Berel-Cantelli Lemma, $\inf\limits_{\w, \bb}l_{p}(\w, \bb|\x, \y,\e)$ converges to $\inf\limits_{\w, \bb}l_{p}(\w, \bb|\x, \y)$ almost surely.

\section{Proof of Lemma \ref{lem:as2}}\label{app:as2}
This lemma is established in NNs with a  single hidden layer and a single output node. By Jensen's Inequality,
\begin{align}
&|\inf\limits_{\w, \bb}l_{p}(\w, \bb|\x,\y)-\inf\limits_{\w, \bb}l_{p}(\w, \bb) |\notag\\
=     &\big|\inf\limits_{\w, \bb}\E_{\e}\left(n^{-1}\textstyle\sum_{i=1}^{n}|f(\x_{i}, \e|\w,\bb)-\y_{i}|^{2}\right)-\notag \\
       & \inf\limits_{\w, \bb}\E_{\e}\left(\E_{\x,\y}|f(\x, \e|\w,\bb)-\y|^2\right)\big|\notag\\
\leq&\;\E_{\e}\big|\inf\limits_{\w, \bb}n^{-1}\textstyle\sum_{i=1}^{n}|f(\x_{i},\e_i|\w,\bb)-\y_{i}|^2-\notag\\
& \quad\inf\limits_{\w, \bb}\E_{\x,\y}|f(\x,\e|\w,\bb)-\y|^2\big|\notag\\
=&\E_{\e}\!\left|\inf\limits_{\w, \bb}l_{p}(\w, \bb|\x,\y,\e)\!-\!\inf\limits_{\w, \bb}l_{p}(\w, \bb|\e)\right|\notag\\
\leq& \E_{\e}\!\left(2\sup\limits_{\w, \bb} |l_{p}(\w, \bb|\x,\y,\e)-l_{p}(\w, \bb|\x,\y,\e)|\right).\label{eqn:jensen}
\end{align}
Define a sequence of functions $\mathcal{F}_{1},\mathcal{F}_{2},\ldots$ as
\begin{align*}
\mathcal{F}_{n}=&\left\{\textstyle\sum_{j=1}^{m_n}w_j^{(2)}f^{(1)}(\x \w_j^{(1)}+b_j^{(1)})+b^{(2)};\right.\\
&\left.\textstyle\w_j^{(1)}\in \mathcal{R}^p, b^{(1)}, b^{(2)}\in \mathcal{R}, \sum_{j=1}^{m_n}|w_j^{(2)}|\leq \beta_n \right\},
\end{align*}
where the number of hidden nodes $m_n$ can change with $n$, and the hidden nodes $f^{(1)}(\x \w_j^{(1)}+b_j^{(1)})$ are uniformly bounded. By Theorem 3 in \citet{Lugosi95}, we have
\begin{align}
\!\!\!\!\!\!\!\!\!\!&\!\!\!\!\!\Pr\!\left(\!\!\sup\limits_{f\in \mathbf{F}_{n}}\!\big|\E_{\x,\y}|f(\x,\e)\!-\!\y|^2\!-\!
\frac{1}{n}\!\sum_{i=1}^{n}|f(\x_i, \e_i)\!-\!\y_i|^2 \big|\!>\!t\! \right)\notag\\
&=\Pr\left(\sup\limits_{f\in \mathbf{F}_{n}}\big|l_{p}(\w, \bb|\e)-l_{p}(\w, \bb|\x,\y,\e) \big|>t \right)\notag\\
& \leq\!4(256e(m_{n}\!+\!1)\beta_n^2/t)^{m_n(2p+3)+1}\exp\!\left(\!\frac{-nt^2}{2048\beta_n^4}\!\right)\!,\!\!\label{eqn:ineq2}
\end{align}
which goes to zero if $n^{-1}m_n\beta_n^4\log(m_n\beta_n)\rightarrow0$, so does
$$\Pr\left(\big| \inf\limits_{\w, \bb}l_{p}(\w, \bb|\x,\y)-\inf\limits_{\w, \bb}l_{p}(\w, \bb) \big|>t \right)$$ by  Eqn  (\ref{eqn:jensen}). If $\exists\;\delta>0$ such that $\beta_n^4/n^{1-\delta}\rightarrow0 $, then $|\inf\limits_{\w, \bb}l_{p}(\w, \bb|\x, \y)-\inf\limits_{\w, \bb}l_{p}(\w, \bb)|<\delta\mbox{ as } n\rightarrow\infty\mbox{ with probability 1}$ is guaranteed by the Borel Borel-Cantelli Lemma.

\section{Proof of Theorem \ref{lem:as4}}\label{app:lem4}
Let $\sigma_{\mbox{\footnotesize{max}}}(n)$  be the maximum noise variance among all injected noises. If $\sigma_{\mbox{\footnotesize{max}}}(n)\rightarrow0$ as $n\rightarrow\infty $, then by  Eqn (\ref{eqn:ineq2}) in Appendix \ref{app:as2}, we have
\begin{align*}
& \Pr\!\left(\!\sup\limits_{f\in \mathbf{F}_{n}}\!\big|l_{p}(\w, \bb|\e)\!-\!l_{p}(\w, \bb|\x,\y,\e) \big|\!>\!t\!\right)\rightarrow\\
&\!\Pr\!\left(\!\sup\limits_{f\in \mathbf{F}_{n}}\!\big|l(\w, \bb)\!-\!l(\w, \bb|\x,\y) \big|\!>t \!\right)\rightarrow\!0
\end{align*}
Taken together with Eqn (\ref{eqn:ineq1}) in Appendix \ref{app:as1}, we have
\begin{equation}\label{eqn:prob4}
\Pr\left(\sup\limits_{f\in \mathbf{F}_{n}}\big|l(\w, \bb)-l_{p}(\w, \bb|\x,\y,\e) \big|>t\right)\rightarrow 0
\end{equation}
as $k\rightarrow\infty, n\rightarrow\infty$. Therefore,   If $\sigma_{\mbox{\footnotesize{max}}}(n)\rightarrow0 \mbox{ as }n\rightarrow\infty $, then $|\inf\limits_{\w, \bb}l_p(\w, \bb|\x,\y,\e)- \inf\limits_{\w, \bb}l(\w, \bb)|<\delta \mbox{ as } k\rightarrow\infty, n\rightarrow\infty$ for any $\delta>0$ with probability 1.

\section{Proof of Theorem \ref{thm:arg}}\label{app:arg}
This theorem is similarly proved as Theorem 3 in \citet{Holmstrom1992}. Suppose Theorem \ref{thm:arg} does not hold for the minimizer $ \hat{\w}_p^{r,n}$ , then there exists an $\epsilon>0$ and a subsequence of $\{n_i\}$ and for each $i$, a subsequence of $(r_{i,j})_j$, such that $d(\nu_{i,j}, \hat{\W}_p^r)\geq \epsilon$ if $\nu_{i,j}=\hat{\w}_p^{r_{i,j},n_i}$ for $i,j\in \N$ (the natural number set). Let $\mu_{i,j}=l_p(\nu_{i,j},\bb|\x,\y,\e)$. By Eqn (\ref{eqn:prob4}), there exists subsequences $(i_k)_k \mbox{ and } (j_k)_k$, such that
\begin{equation}\label{eqn:app5}
\Pr\!\left(\!\sup\limits_{\w} |\mu_{i_k,j_k}(\w) -l(\w, \bb)|\! >\! t\!\right)\!\leq \! k^{-1}, k\in \N.\!
\end{equation}
Since $\W$ is compact, the subsequence $(\nu_{i_k,j_k})_k$ converges to a point $\hat{\w}^*\in \W$ and $\hat{\w}^*\notin \hat{\W}^0$  since $d( \hat{\w}^*, \hat{\W}^0)\! \geq \epsilon$. On the other hand, for an arbitrary $\w\in \W$, we have
\begin{align*}
& l( \hat{\w}^*,\bb)-l(\w,\bb)]=\\
&(l( \hat{\w}^*,\bb)-l(\nu_{i_k,j_k},\bb))+(l( \nu_{i_k,j_k},\bb)-\mu_{i_k,j_k}(\nu_{i_k,j_k}) )\\
&\quad\!(\mu_{i_k,j_k}(\nu_{i_k,j_k})-\mu_{i_k,j_k}(\w))+(\mu_{i_k,j_k}(\w)-l(\w,\bb))
\end{align*}
By the continuity of the ilf and $\lim\limits_{j\rightarrow\infty}\nu_{i_k,j_k}=\hat{\w}^*$, $l( \hat{\w}^*,\bb)-l(\nu_{i_k,j_k},\bb)$ in the above equation is arbitrarily small with $k\rightarrow\infty$; by Eqn (\ref{eqn:app5}), $l( \nu_{i_k,j_k},\bb)-\mu_{i_k,j_k}(\nu_{i_k,j_k}) $ and $(\mu_{i_k,j_k}(\w)-l(\w,\bb))$ are arbitrarily small with $k\rightarrow\infty$; and $(\mu_{i_k,j_k}(\nu_{i_k,j_k})-\mu_{i_k,j_k}(\w))$ is non-positive. By the arbitrariness of $\w \in \W$, we must have $\hat{\w}^*\in \W^0$, which is a contradiction to the assumption that Theorem \ref{thm:arg} fails.

\section{Proof of Corollary \ref{cor:tail}}\label{app:tail}
The proof utilize the following theorem \citep{Grandvalet1997}. Let $\boldsymbol{\epsilon}=(\epsilon_1, \ldots,\epsilon_p)$ be a vector of $p$ independent standard Gaussian variables and $g\!:\!R^{p}\!\rightarrow\!R$ be $L$-Lipschitz continuous with respect to the Euclidean norm, then $g(\boldsymbol{\epsilon})-\mbox{E}(g(\boldsymbol{\epsilon}))$ is sub-Gaussian with parameter at most $L$ such as, for any $\delta\geq0$,
\begin{equation}\label{eqn:subgaussian1}
\Pr(| g(\boldsymbol{\epsilon})-E(g(\boldsymbol{\epsilon}))| > \delta)\leq2\exp\left(\frac{- \delta^{2}}{2L^{2}}\right).
\end{equation}
In the context of whiteout, injected noise $\e$ has mean $\mathbf{0}$ and covariance $\Sigma\!=\!\mbox{diag}\left\{\sigma^2|\w|^{-\gamma}+\lambda\right\}$. We re-write $\e=\Sigma^{1/2}\boldsymbol{\epsilon}$, where $\boldsymbol{\epsilon}$ is standard Gaussian with mean $\mathbf{0}$ and the identity covariance. $g(\boldsymbol{\epsilon})$ in Eqn (\ref{eqn:subgaussian1}) refers to pelf $l_p(\w, \bb|\x,\y,\e)\!\!=\!\!
(kn)^{-1}\sum_{j=1}^{k}\!\sum_{i=1}^{n}\!(f(\x_{i},\Sigma_{ij}^{1/2}\boldsymbol{\epsilon}|\w,\bb)\!-\!y_{i})^{2}$, and $E(g(\boldsymbol{\epsilon}))$ is nm-pelf $l_p(\w, \bb|\x,\y)$ in the whiteout setting, where $f$ is the composition of a series of  continuous and bounded activation functions between layers. To determine the Lipschitz constant for $g(\boldsymbol{\epsilon})$ in the context of whiteout, we need to bound $\frac{\partial g}{\partial \boldsymbol{\epsilon}}$. Applying the chain rule, we have $\frac{\partial g}{\partial \boldsymbol{\epsilon}}=2(kn)^{-1}\sum_{j=1}^{k}\!\sum_{i=1}^{n}\!(f(\x_{i},\Sigma_{ij}^{1/2}\boldsymbol{\epsilon}|\w,\bb)-y_{i})
f'(\x_{i},\Sigma_{ij}^{1/2}\boldsymbol{\epsilon}|\w,\bb)\w$. $f$ and $\y$ are bounded;  the derivative of $f$ is bounded per its Lipschitz continuity condition,  and $\w$ is naturally bounded due to the corresponding regularization constraints imposed by whiteout, $l_p(\w, \bb|\x,\y,\boldsymbol{\epsilon})$ is  therefore Lipschitz continuous. Since $\e$ is a linear function of $\boldsymbol{\epsilon}$, the Lipschitz continuity maintains in terms of $\e$. 
Then by the Bounded differences inequality theorem \citep{wainwright2018}, the the Lipschitz constant of $l_p(\w, \bb|\x,\y,\e)$ is $B/\sqrt{kn}$, where $B>0$ is a constant, then
$$\Pr(| l_p(\w, \bb|\x,\y,\e)\!-\!l_p(\w, \bb|\x,\y) |\!>\! \delta)
\!\leq\! 2\exp\!\left(\!-\frac{kn\delta^{2}}{2B^2}\right)$$
\section{Backpropagation with Without}\label{sec:bp}
The backpropagation (BP) algorithm often used training NNs in the supervised learning setting  comprises a feedforward (FF) step and a BP step. We list a revised BP algorithm in Table \ref{tab:BPsteps} to accommodate whiteout NI using feedforward NNs an example.  In brief, whiteout affects the calculation of hidden units in the FF step and the gradients of the loss function  in the BP step, while all the other steps remain unchanged compared to the regular BP algorithm.

Let $m^{(l)}$ and $m^{(l+1)}$  denote the number of nodes in layer $l$ and $l+1$, respectively, where $l\!=\!1,\ldots,L\!-\!1$ and $L$ is the number of fully-connected layers.   The training loss $D$ measures the difference between the observed outcomes in the training data  and their predicted values from the learned NN. The BP  steps  in Table \ref{tab:BPsteps} are illustrated using the additive whiteout noise (Eqn \ref{eqn:additive}); and the steps with the multiplicative noise (Eqn \ref{eqn:multiplicative}) are listed in the footnote of Table \ref{tab:BPsteps}.

\begin{table}[!htp]
\resizebox{\linewidth}{!}{
\begin{tabular}{l}
\hline
\textbf{input}:  learning rate $\eta$; tuning  parameters in whiteout Gaussian \\
 noise $(\sigma^2, \gamma, \lambda)$; initial values for $\w$ and $\bb$. \\
1. \textbf{FF}:\\
    \hspace{8pt} \textbf{do} $l=1 \mbox{ to } L-1$ \\
    \hspace{14pt} \textbf{do} $k=1 \mbox{ to } m^{(l+1)}$ \\
	 \hspace{14pt} $\bullet$ sample $e_{1k}, \ldots, e_{m^{(l)}k}$ from $N(0, 1)$:\\
	 \hspace{14pt} $\bullet$  calculate $u_k^{(l+1)}\!\!=\!b_k^{(l)}\!\!+\!\!\sum_{j=1}^{m^{(l)}}\!w_{jk}^{(l)}\left(\!\!X_j^{(l)}\!+\!e_{jk}\sqrt{\sigma^{2}|w_{jk}^{(l)}|^{-\gamma}\!+\!\lambda}\right)\!$ \\
 \hspace{24pt} and  $X_k^{(l+1)}\!=\!f\!\left(\!u_k^{(l+1)}\!\right)$.\\
    \hspace{14pt} \textbf{end do}\\
     \hspace{8pt} \textbf{end do}\\
2. \textbf{BP}:\\
    \hspace{8pt} \textbf{do} $l=L-1 \mbox{ to } 1$ \\
    \hspace{14pt} \textbf{do} $k=1 \mbox{ to } m^{(l+1)}$ \\
    \hspace{20pt} \textbf{do} $j=1 \mbox{ to } m^{(l)}$ \\
    \hspace{20pt} $\bullet$   $^\S$update weight: $w_{jk}^{(l)}=w_{jk}^{(l)}-\eta \frac{\partial D}{\partial w_{jk}^{(l)}}$, where\\
	\hspace{28pt}  $\frac{\partial D}{\partial w_{jk}^{(l)}}=\frac{\partial D}{\partial u_k^{(l+1)}}\frac{\partial u_k^{(l+1)}}{\partial w_{jk}^{(l)}}$ with $\frac{\partial D}{\partial u_k^{(l+1)}}\!=\!\frac{\partial D}{\partial X_k^{(l+1)}}f'\!\left(u_k^{(l+1)}\right)$ \\
\hspace{28pt} and $\frac{\partial u_{k}^{(l+1)}}{\partial w_{jk}^{(l)}}\!=\!X_j^{(l)}+e_{jk}\frac{2(\sigma^{2}| w_{jk}^{(l)}|^{-\gamma}+\lambda)- \sigma^2\gamma|w_{jk}^{(l)}|^{-\gamma}}{2\left(\sigma^2| w_{jk}^{(l)}|^{-\gamma}+\lambda\right)^{-1/2}}$ \\
\hspace{20pt} \textbf{end do}\\
    \hspace{20pt} $\bullet$   update bias: $b_k^{(l)}=b_k^{(l)}-\eta \frac{\partial D}{\partial b_k^{(l)}}$, where\\
	\hspace{28pt} $\frac{\partial D}{\partial b_k^{(l)}}=\frac{\partial D}{\partial u_k^{(l+1)}}\frac{\partial u_k^{(l+1)}}{\partial b_k^{(l)}}=\frac{\partial D}{\partial u_k^{(l+1)}}$\\
    \hspace{14pt} \textbf{end do}\\
     \hspace{8pt} \textbf{end do}\\
    \textbf{Repeat} the FF and BP steps for a sufficient number of epochs\\
        until the training loss $D$ converges. \\
    \textbf{output}: estimates of $(\w,\bb)$.\\
	\hline
\footnotesize{$^\S$e.g.: if $D\!=\!\frac{1}{2}\sum_{i=1}^{n}(y_i-\hat{y}_i)^{2}$, then $\frac{\partial D}{\partial u_k^{(L)}}\! =\! \! \left(Y\!-\!X_k^{(L)}\right)f'\!\left(\!u_k^{(L)}\!\right)$ for} \\
\footnotesize{ $l\!=\!L\!-\!1$;  $\frac{\partial D}{\partial u_k^{(L-1)}}\! =\!\frac{\partial D}{\partial X_k^{(L-1)}}f'\!\left(u_k^{(L-1)}\right)
=\sum_{k'}\frac{\partial D}{\partial u_{k'}^{(L)}}\frac{\partial u_{k'}^{(L)}}{\partial X_{k}^{(L-1)}}f'\!\left(u_k^{(L-1)}\right)$}\\
\footnotesize{$=\sum_{k'}\frac{\partial D}{\partial u_{k'}^{(L)}} w_{kk'}^{(L-1)}f'\!\left(\!u_k^{(L-1)}\!\right)$ for $l=L-2$, and so on for $l\!=\!L-3,\ldots,1$}\\
\small{All the steps above are the same with the multiplicative noise, except for} \\
\small{the calculation of  $u_k^{(l+1)}$ in FF  and  $\frac{\partial u_k^{(l+1)}}{\partial w_{jk}^{(l)}}$ in BP.} \\
\footnotesize{\textbf{FF: }  $u_k^{(l+1)}=b_k^{(l)}+ \sum_{j=1}^{m^{(l)}}w_{jk}^{(l)}X_j^{(l)}\left(1+e_{jk}\sqrt{\sigma^{2}|w^{(l)}_{jk}|^{-\gamma}+\lambda}\right)$;}\\
\footnotesize{\textbf{BP: } $\frac{\partial u_j^{(l+1)}}{\partial w_{jk}^{(l)}}=X_j^{(l)}+ X_j^{(l)}e_{jk}\frac{2(\sigma^{2}| w_{jk}^{(l)}|^{-\gamma}+\lambda)- \sigma^2\gamma|w_{jk}^{(l)}|^{-\gamma}}{2\left(\sigma^2| w_{jk}^{(l)}|^{-\gamma}+\lambda\right)^{-1/2}}$.}\\
\hline
\end{tabular}}
\caption{Backpropagation with Whiteout NI}\label{tab:BPsteps}
\end{table}


\section{The NN structure employed in the MNIST and CIFAR-10 experiments}\label{sec:NNstructure}
In both structures, Layer 1 and layer 2 were convolutional layers followed by ReLU nonlinear activation and max-pooling; layer 3 and layer 4 were fully-connected layers.
\begin{table}[!htp]
\centering\resizebox{\linewidth}{!}{
\begin{tabular}{l|c|c|c|c}
\hline
Layer & 1 & 2 & 3 & 4 \\\hline
Type &conv. &conv.&$\!\!$fully-conn.$\!\!$&$\!\!$fully-conn.$\!\!$\\
Channels/Nodes & 20 & 50 &500 & 10 \\
Filter Size & $5\times5$ & $5\times5$ & - & - \\
Conv. Stride & 1 & 1 & - & - \\
Pooling Size & $2\times2$ & $2\times2$ & - & - \\
Pooling Stride & 2 & 2 & - & - \\
Activation &$\!\!$ReLU$\!\!$&$\!\!$ReLU$\!\!$&ReLU&Softmax\\
\hline
\end{tabular}}
\caption{The NN structure employed in the MNIST data (reproduced from \citet{Kang2016})}\label{tab:mnist1}
\end{table}

\begin{table}[!htp]
\centering\resizebox{\linewidth}{!}{
\begin{tabular}{l|c|c|c|c|c}
\hline
Layer & 1 & 2 & 3 & 4 & 5 \\\hline
Type &conv. &conv. &conv. &$\!\!$fully-$\!\!$&$\!\!$fully-$\!\!$\\		
&&&&conn.&conn.\\
Nodes & 32 & 32 &64 & 64 & 10 \\
Filter Size &$\!\!\!5\times5\!\!\!$& $5\times5$ & $5\times5$ & - & - \\
Conv. Stride & 1 & 1 & 1 & - & - \\
Pooling type & max &$\!\!\!\!$average$\!\!\!\!$&$\!\!\!\!$average$\!\!\!\!$& - & - \\
Pooling Size &$\!\!\!3\times3\!\!\!$& $3\times3$ & $3\times3$ & - & - \\
Pooling Stride$\!\!\!$& 2 & 2 & 2 & - & - \\
Activation &$\!\!$ReLU$\!\!$&ReLU&ReLU&ReLU&Softmax\\
\hline
\end{tabular}}
\caption{The NN structure employed in the CIFAR-10 data (reproduced from \citet{Kang2016})}\label{tab:cifar1}
\end{table}

\section{Example empirical distributions of the learned weights in the LSTV and LIBRAS experiments}\label{sec:density}

\begin{figure}[!htb]\begin{center}
\includegraphics[ height=0.7\linewidth]{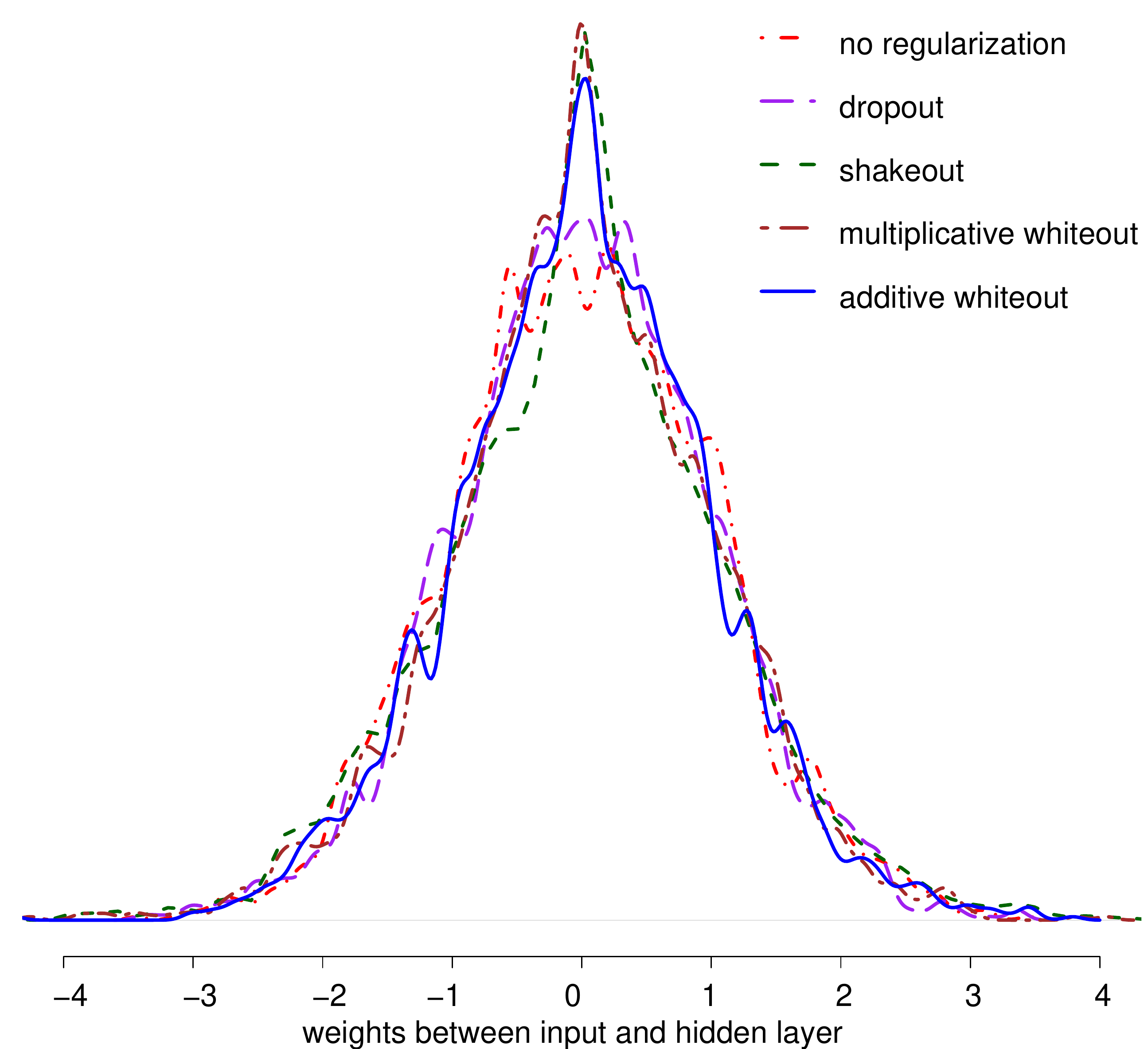}
\caption{Example empirical distributions of the learned weights connecting the  NN input and hidden layer  in the LSTV data}\label{fig:pdfLSVT}
\end{center}\end{figure}

\begin{figure}[!htp]\begin{center}
\includegraphics[width=0.75\linewidth]{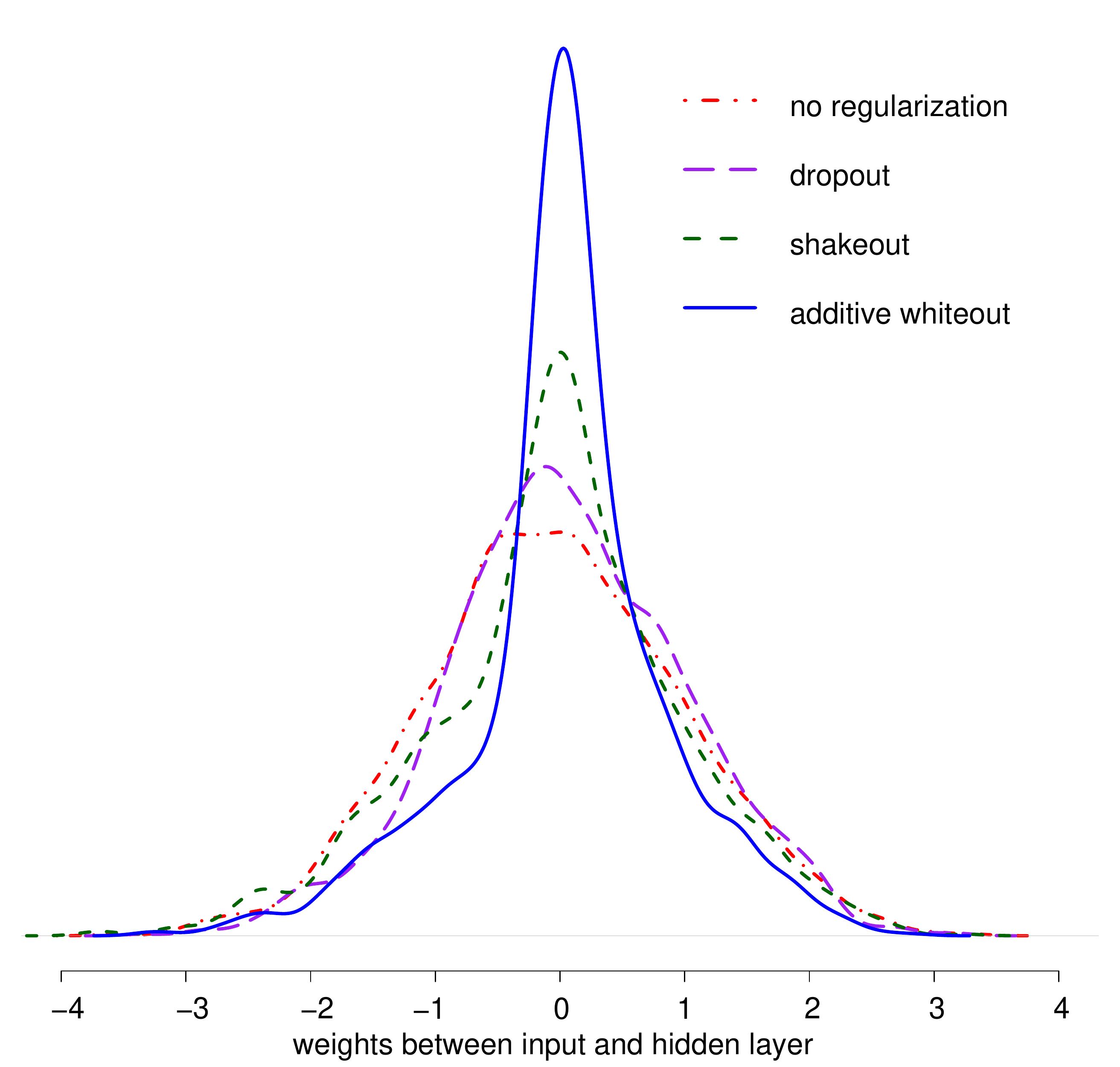}
\caption{Example empirical distributions of the learned weights connecting the input and the first hidden layer (training size = 60) in the LIBRAS data}\label{fig:pdfmove}\end{center}
\end{figure}

\section{Whiteout and NIRTs in Unsupervised Learning}\label{sec:unsupervised}
Whiteout, or NI in general, can also be applied to unsupervised learning  such as dimension reduction and pre-training of deep NNs. Here we illustrate the applications of whiteout in RBMs (Section \ref{sec:rbm}) and auto-encoders (AE)  (Section \ref{sec:ae}), both of which are widely  used unsupervised learning methods. The RBM algorithm is known for providing good initial values for the weights in a deep NN to facilitate later supervised learning via the NN, and AEs can be used for dimensionality reduction and learning generative models of data.  

\subsection{RBM}\label{sec:rbm}
The main difference in training a RBM with NI vs. without is the form of the conditional distribution of the visible nodes given the hidden nodes in a RBM and vise versa; and the rest of the procedures are the same with or without NI. Consider a RBM with visible nodes $\vv\in \{0,1\}^p$ and hidden nodes $\h\in \{0,1\}^m$. The joint probability distribution of $\h$ and $\vv$ is  $p(\h,\vv|\w,\aaa,\bb)= \exp(-F(\vv,\h))(A(\w,\aaa,\bb))^{-1}$,
where $F(\vv,\h)\!=\!-(\vv^{t}\w\h+\aaa^{t}\h+\bb^{t}\vv)$ is the free energy, $A=A(\w,\aaa,\bb)\!=\!\sum_{\vv}\sum_{\h}\exp\{-F(\vv,\h)\} $ is the normalization constant,  $\w\!=\!\{w_{ij}\}\!\in\!\mathcal{R}^{p\times m}$ is the weight matrix connecting the visible and the hidden layers, and $\aaa$ and $\bb$ are the bias parameters. The parameters are trained by maximizing the likelihood of the parameters given the observed $\vv$ over $n$ observations,
$ L(\w,\aaa,\bb|\vv)=\textstyle\prod_{i=1}^{n}\sum_{\h_i}p(\h_i,\vv_i|\w,\aaa,\bb)= \prod_{i=1}^{n} \sum_{\h_i}\exp(-\F(\vv_i,\h_i))(A(\w,\aaa,\bb))^{-1}$.

Whiteout or NIRTs in general may regulate the RBM with noise $\e\in R^{p\times m}$ injected into the $p$ input nodes during the training in one epoch. Denote the noise  perturbed inputs by $\tilde{\vv}$, the joint distribution of $(\h,\tilde{\vv})$ is
$$p(\h,\tilde{\vv}|\e,\w,\aaa,\bb)= \exp(-F(\tilde{\vv},\h))(\tilde{A}(\w,\aaa,\bb))^{-1},$$
where  $\tilde{A}(\w,\aaa,\bb)=\sum_{\tilde{\vv}}\sum_{\h}\exp\{-F(\tilde{\vv},\h)\} $, and the corresponding likelihood  is
\begin{equation}\label{eqn:marg.likelihood}
\!\!\!\!L_p(\w,\aaa,\bb|\tilde{\vv})\!=\!\textstyle\prod_{i=1}^{n}\! \sum_{\h_i}\!e^{-F(\tilde{\vv}_i,\h_i)}(\tilde{A}(\w,\aaa,\bb))^{-1}.\!\!
\end{equation}
In expectation with regard to noise, the noise-injected loss function $\sum_{i=1}^{n}l_p(\w,\aaa,\bb|\tilde{\vv}_i)$ is the corresponding negative log-likelihood $-\log(L_p(\w,\aaa,\bb|\tilde{\vv}))$
\begin{align}\label{eqn:rbm}
\approx&\textstyle \sum_{i=1}^{n}l(\w,\aaa,\bb|\vv_i) +\\
&\textstyle\frac{1}{2}\! \sum_{i=1}^{n}\!\sum_{k=1}^p\!\left\{ \mbox{V}(e_{ik})\!
\!\left(\!\mbox{V}_{\vv,\h}\!\left(\!\sum_{j=1}^{m}\!w_{kj}h_{ij}\!+\!b_k\!\right)\right.\right.\notag\\
&\textstyle\qquad\qquad\qquad\left.\left.-\mbox{V}_{\h|\vv}\!\left(\!\sum_{j=1}^{m}\!w_{kj}h_{ij}\!+\!b_k\!\right)\!\right)\!\right\},\notag
\end{align}
where  $\sum_{i=1}^{n}l(\w,\aaa,\bb|\vv_i)$ is the loss function (negative log-likelihood) with the original data, $\mbox{V}_{*}$ is the variance operator with regard to the distribution of $*$.  Eqn (\ref{eqn:rbm}) suggests that minimizing the noise-injected loss function with whiteout in RBM is approximately equivalent, in expectation with respect to $\e$, to minimizing the original loss function, plus a penalty terms on the RBM parameters. The proof of Eqn (\ref{eqn:rbm}) is given below. Based on Eqn (\ref{eqn:marg.likelihood}),
\begin{align} \label{eqn:app.rbm}
&\textstyle\sum_{i=1}^{n}l_p(\w,\aaa,\bb|\tilde{\vv}_i)\notag\\
\!=&\textstyle\sum_{i=1}^{n} \left(\!\log\left(\!\sum_{\tilde{\vv_i},\h_i}\!e^{-F(\tilde{\vv}_i,\h_i)}\right)\!
  \!-\! \log\!\left(\sum_{\h_i}\!e^{-F(\tilde{\vv}_i,\h_i)}\right)\!\right)\notag\\
&\textstyle\E_\e\left(\sum_{i=1}^{n}l_p(\w,\aaa,\bb|\tilde{\vv}_i)\right)\\
&\!\!=\!\!\sum_{i=1}^{n}\E_\e\!\!\left(\!\underbrace{\log\!\left(\!\sum_{\tilde{\vv}_i,\h_i}\!e^{-F(\tilde{\vv}_i,\h_i)}\!\right)}_{\text{term 1}}
\!  \!-\! \underbrace{\log\!\left(\!\sum_{\h_i}\!e^{-F(\tilde{\vv}_i,\h_i)}\!\right)}_{\text{term 2}}\!\right),\notag
\end{align}
\normalsize where $F(\vv,\h)=\vv^{t}\w\h+\aaa^{t}\h+\bb^{t}\vv$. Let $B=B(\vv,\w,\aaa,\bb)= \sum_{\h}\exp(-F(\vv,\h))$, and apply the second-order Taylor expansion  around $m$ repeats of $\vv_i$ on terms 1  and 2, respectively, we have
\begin{align*}
&\mbox{\textbf{term 1}}\!\approx\!\log(A)\!- \!A ^{-1}\!\textstyle\left(\sum_{\vv_i,\h_i}e^{-F(\vv_i,\h_i)}\mathbf{g}(\vv_i,\h_i)\right)\e_i\\
&\!+\!\textstyle\frac{1}{2}A^{-1} \e'_i\left(\!\sum_{\vv_i,\h_i}e^{-F(\vv_i,\h_i)}\mathbf{H}(\vv_i,\h_i) -\right.\\
&\quad\quad\textstyle\left.\sum_{\vv_i,\h_i}e^{-F(\vv_i,\h_i)}\mathbf{g}(\vv_i,\h_i)\mathbf{g}'(\vv_i,\h_i)\!\right)\!\e_i\\
&-\textstyle\frac{1}{2}A^{-2}\e_i'\left(\textstyle\sum_{\vv_i,\h_i}e^{-F(\vv_i,\h_i)}\mathbf{g}(\vv_i,\h_i)\right)\\
&\quad\quad \left(\textstyle\sum_{\vv_i,\h_i}e^{-F(\vv_i,\h_i)}\mathbf{g}(\vv_i,\h_i)\right)'\e_i,\\
&\mbox{\textbf{term 2}}\approx\\ &\log(B(\vv_i))\!-\!(B(\vv_i))^{-1}\textstyle\!\left(\sum_{\h_i}\!e^{-F(\vv_i,\h_i)}\mathbf{g}(\vv_i,\h_i)\right)\!\e_i\\
&\!+\!\textstyle\frac{1}{2}(B(\vv_i))^{-1}\e_i'\!\left(\sum_{\h_i}\!e^{-F(\vv_i,\h_i)}\mathbf{H}(\vv_i,\h_i)-\right.\\
&\quad\quad\left.\textstyle\sum_{\h_i}\!e^{-F(\vv_i,\h_i)}\mathbf{g}(\vv_i,\h_i)\mathbf{g}'(\vv_i,\h_i)\right)\!\e_i\\
&-\textstyle\frac{1}{2}(B(\vv_i))^{-2}\e_i'\left(\sum_{\h_i}e^{-F(\vv_i,\h_i)}\mathbf{g}(\vv_i,\h_i)\right)\\
&\quad\quad\left(\textstyle\sum_{\h_i}e^{-F(\vv_i,\h_i)}\mathbf{g}(\vv_i,\h_i)\right)'\e_i,
\end{align*}
where $\e_i$ contains the independent noises added to $p$ visible nodes $\vv_i$, $\mathbf{g}(\vv_i,\h_i)$ of dimension $mp\times1$ contains the gradients of $F(\vv_i,\h_i)$ with regard to $m$ repeats of $\vv_i$, and $\mathbf{H}(\vv_i,\h_i)$ of dimension $mp\times mp$ is the Hessian matrix of $F(\vv_i,\h_i)$ with regard to $m$ repeats of $\vv_i$. Therefore, Eqn (\ref{eqn:app.rbm}) is approximately
\begin{align*}
&\textstyle\sum_{i=1}^{n}l_p(\w,\aaa,\bb|\tilde{\vv}_i)\approx\textstyle\sum_{i=1}^{n}l(\w,\aaa,\bb|\vv_i)+\\
&\textstyle\frac{1}{2}\sum_{i=1}^{n} \E\left[ \e'_i\colorbox{gray!50}{$\textstyle A^{-1}\left(\!\sum_{\vv_i,\h_i}e^{-F(\vv_i,\h_i)}\mathbf{H}(\vv_i,\h_i)-\right.$}\right.\\
&\qquad\qquad\qquad\textstyle\colorbox{gray!50}{$\left.\sum_{\vv_i,\h_i}e^{-F(\vv_i,\h_i)}\mathbf{g}(\vv_i,\h_i)\mathbf{g}'(\vv_i,\h_i)\!\right)\!$}\e_i\\
&- \e_i'\colorbox{gray!50}{$\textstyle A^{-2}\left(\textstyle\sum_{\vv_i,\h_i}e^{-F(\vv_i,\h_i)}\mathbf{g}(\vv_i,\h_i)\right)$}\\
&\qquad\colorbox{gray!50}{$\textstyle\left(\sum_{\vv_i,\h_i}e^{-F(\vv_i,\h_i)}\mathbf{g}(\vv_i,\h_i)\right)'$}\e_i\\
&\!-\!\e_i'\colorbox{gray!50}{$\textstyle (B(\vv_i))^{-1}\!\!\left(\sum_{\h_i}\!e^{-F(\vv_i,\h_i)}\mathbf{H}(\vv_i,\h_i)\right.$}-\\
&\qquad\textstyle\colorbox{gray!50}{$\left.\sum_{\h_i}\!e^{-F(\vv_i,\h_i)}\mathbf{g}(\vv_i,\h_i)\mathbf{g}'(\vv_i,\h_i)\right)\!$}\e_i\\
& +\e_i'\colorbox{gray!50}{$\textstyle (B(\vv_i))^{-2}\left(\sum_{\h_i}e^{-F(\vv_i,\h_i)}\mathbf{g}(\vv_i,\h_i)\right)$}\\
&\left.\qquad\textstyle\colorbox{gray!50}{$\left(\sum_{\h_i}e^{-F(\vv_i,\h_i)}\mathbf{g}(\vv_i,\h_i)\right)'\!$}\e_i\right],\\
&\!=\!\textstyle\sum_{i=1}^{n}l(\w,\aaa,\bb|\vv_i)\! +\!\textstyle\frac{1}{2}\textstyle \sum_{i=1}^{n}
\E\left[ \e_i'\mathbf{U}_i\e_i\right]\!,\\
&\mbox{where $\mathbf{U}_i$ is the sum of the $\!\!$\colorbox{gray!50}{highlighted terms}}\!
\end{align*}
Since $F$ is linear in $\vv_i$, so $H$ is matrix of 0, and thus $\mathbf{U}_i=$
\begin{align*}
&\textstyle(B(\vv_i))^{-1}\left(\sum_{\h_i}e^{-F(\vv_i,\h_i)}\mathbf{g}(\vv_i,\h_i)\mathbf{g}'(\vv_i,\h_i)\right)+\\
&\textstyle\left((B(\vv_i))^{-1}\sum_{\h_i}e^{-F(\vv_i,\h_i)}\mathbf{g}(\vv_i,\h_i)\right)\\
&\textstyle\left((B(\vv_i))^{-1}\sum_{\h_i}e^{-F(\vv_i,\h_i)}\mathbf{g}(\vv_i,\h_i)\right)'-\\
&\textstyle(A^{-1}\left(\sum_{\vv_i,\h_i}e^{-F(\vv_i,\h_i)}\mathbf{g}(\vv_i,\h_i)\mathbf{g}'(\vv_i,\h_i)\right)-\\
&\textstyle\left(A^{-1}\sum_{\vv_i,\h_i}e^{-F(\vv_i,\h_i)}\mathbf{g}(\vv_i,\h_i)\right)\\
&\textstyle\left(A^{-1}\sum_{\vv_i,\h_i}e^{-F(\vv_i,\h_i)}\mathbf{g}(\vv_i,\h_i)\right)'\\
=&\;\textstyle\E_{\vv,\h}\!\left(\mathbf{g}\mathbf{g}'\right)+\left(\E_{\vv_i,\h_i}(\mathbf{g}_i)\right)\left(\E_{\vv,\h}(\mathbf{g}_i)\right)'-\\
&\;\textstyle\E_{\h|\vv}\!\left(\mathbf{g}_i\mathbf{g}_i'\right)+\left(\E_{\h_i|\vv_i}(\mathbf{g}_i)\right)\left(\E_{\vv_i,\h_i}(\mathbf{g}_i)\right)'\\
= &\textstyle \mbox{V}_{\vv_i,\h_i}(\mathbf{g}_i)- \mbox{V}_{\h_i|\vv_i}(\mathbf{g}_i)
\end{align*}

\subsection{AE}\label{sec:ae}
In the case of an AE,  the goal is often to reconstruct the inputs from a learned AE. NI into input nodes has been shown to help achieve better generalization results in denoising AEs \citep{dae}. Empirical results of AEs with dropout and shakeout noise injected into hidden nodes are given in \citet{dropout14} and \citet{Kang2016}.  Here we briefly illustrate the theoretical motivation underlying NI in an AE, focusing on the parameters (weights) in the decoder part of the AE (results are generalized automatically to the weights in the encoder  with the often used tied-weights assumption). Since an AE is trained layerwise, it is sufficient to illustrate the idea using a single pair of layers.  Let $g(\cdot)$ denote the activation function between the hidden nodes $\h_i=(h_{i1}, \ldots,h_{im})$  and the output nodes  $\x_i=(x_{i1}, \ldots,x_{ip})^t$ for cases $i=1,\ldots,n$. Empirical loss is defined by $l(\w,\bb|\x)\!=\!\sum_{i=1}^{n}\!\sum_{j=1}^{p}(x_{ij}-g(\h_i\w_{.j} +b_j))^2$, where $\w\!=\!\{w_{kj}\}_{m\times p}$ and $\bb=(b_1,\ldots,b_p)$ represent the weight and bias parameters between the two layers.  The approximate expected  perturbed loss with noise $\e$ injected into the hidden nodes is a penalized version of the original loss with a penalty term on $\w$,
\begin{align}
\E_{\e}&(l_p(\w,\bb|\x,\e))\approx l(\w,\bb|\x)\!\notag\\
&+\!2\!\textstyle\sum_{i=1}^{n}\!\sum_{j=1}^{p}\left(\sum_{k=1}^{m}w_{kj}^2 \mbox{V}( e_{ijk})\right)\label{eqn:ae}\\
&\textstyle\!\left(\!\left(\frac{\partial g(\h_i\w_{.j} \!+\!b_j)}{\partial h_{ij}}\!\right)^2-\frac{\partial^2 g( \h_i\w_{.j}\!+\!b_j)}{\partial h_{ij}^2}(g( \h_i\w_{.j}\!+\!\bb)\!-\!x_{ij})\!\right).\notag
\end{align}
The proof of Eqn (\ref{eqn:ae}) is given below.
\begin{align*}
& \E_{\e}(l_p(\w,\bb|\x,\e))\\
&=\E_{\e}\left(\textstyle\sum_{i=1}^{n}\!\sum_{j=1}^{p}(x_{ij}-g((\h_i+\e_{ij.})\w_{.j} +b_j))^2\right)\\
&\approx\textstyle\sum_{i=1}^{n}\!\sum_{j=1}^{p}(\x_{ij}-g( \h_i\w_{.j} \!+\!b_j))^2+\\
&\textstyle\!\E_{\e}\!\!\left(\!\sum_{i=1}^{n}\!\sum_{j=1}^{p}\!2(g( \h_i\w_{.j} \!+\!b_j)\!-\!\x_{ij})
\frac{\partial g( \h_i\w_{.j} \!+\!b_j)}{\partial h_{ij}} \e_{ij.}\w_{.j}\!\right)\\
&+\textstyle\E_{\e}\!\left(\!\sum_{i=1}^{n}\!\!\sum_{j=1}^{p}\!\left(\left(\!\frac{\partial g(\h_i\w_{.j} \!+\!b_j)}{\partial h_{ij}}\!\right)^2-\right.\right.\\
&\textstyle\left.\left.\qquad\frac{\partial^2 g(\w_i \h_k\!+\!\bb)}{\partial h_{ij}^2}(g(\h_i\w_{.j} \!+\!b_j)\!-\!x_{ij})\!\right) (\e_{ij.}\w_{.j})^2 \right)\\
=&l(\w,\bb|\x)+\textstyle\sum_{i=1}^{n}\!\!\sum_{j=1}^{p}\!\left(\sum_{k=1}^{m}w_{kj}^2 \mbox{V}( e_{ijk})\right)\\
&\textstyle\left(\!\left(\frac{\partial g(\h_i\w_{.j} \!+\!b_j)}{\partial h_{ij}}\!\right)^2-\frac{\partial^2 g(\h_i\w_{.j} \!+\!b_j)}{\partial h_{ij}^2}(g(\h_i\w_{.j} \!+\!b_j)\!-\!x_{ij})\!\right)
\end{align*}

\end{document}